\documentclass[11pt, a4paper, logo, copyright, numbering]{googledeepmind}

\usepackage[utf8]{inputenc}

\usepackage[authoryear, sort&compress, round]{natbib}

\usepackage{xcolor}
\usepackage{lipsum}
\usepackage{amsmath}
\usepackage{amsfonts}
\usepackage{float}
\usepackage{wrapfig}
\usepackage{tikz}
\usepackage{pgfplots}
\usepackage{subcaption}
\usepackage{wrapfig}
\usepackage{tcolorbox}
\usepackage{array, makecell}
\usepackage{graphicx}
\usepackage{varwidth}
\usetikzlibrary{shapes.callouts, shadows}
\usetikzlibrary{shapes.multipart}
\usepackage{listings} 
\usepackage{lineno}
\usepackage{subcaption}
\usetikzlibrary{arrows.meta, positioning,patterns,arrows}
\usepgfplotslibrary{groupplots}
\usetikzlibrary{backgrounds,calc,bending,backgrounds}
\pgfplotsset{compat=1.16}
\usepackage{pgfplotstable}
\usetikzlibrary{matrix, positioning}
\usepackage[]{mdframed}

\definecolor{chartreuse}{rgb}{0.75, 0.9, 0.7}

\title{BIG-Bench Extra Hard}
\correspondingauthor{mehrankazemi@google.com and baharef@google.com}

\newcommand\blfootnote[1]{%
  \begingroup
  \renewcommand\thefootnote{}\footnote{#1}%
  \addtocounter{footnote}{-1}%
  \endgroup
}

\author[1]{Mehran Kazemi}
\author[2]{Bahare Fatemi}
\author[1,3]{Hritik Bansal}
\author[1]{John Palowitch}
\author[1]{Chrysovalantis Anastasiou}
\author[1]{Sanket Vaibhav Mehta}
\author[2]{Lalit K. Jain}
\author[1]{Virginia Aglietti}
\author[1]{Disha Jindal}
\author[1]{Peter Chen}
\author[2]{Nishanth Dikkala}
\author[1]{Gladys Tyen}
\author[2]{Xin Liu}
\author[1]{Uri Shalit}
\author[1]{Silvia Chiappa}
\author[1]{Kate Olszewska}
\author[1]{Yi Tay}
\author[1]{Vinh Q. Tran}
\author[1]{Quoc V. Le}
\author[1]{Orhan Firat}

\affil[1]{Google DeepMind}
\affil[2]{Google Research}
\affil[3]{UCLA}

\begin{document}

\begin{abstract}
    Large language models (LLMs) are increasingly deployed in everyday applications, demanding robust general reasoning capabilities and diverse reasoning skillset. However, current LLM reasoning benchmarks predominantly focus on mathematical and coding abilities, leaving a gap in evaluating broader reasoning proficiencies. One particular exception is the BIG-Bench dataset, which has served as a crucial benchmark for evaluating the general reasoning capabilities of LLMs, thanks to its diverse set of challenging tasks that allowed for a comprehensive assessment of general reasoning across various skills within a unified framework. However, recent advances in LLMs have led to saturation on BIG-Bench, and its harder version BIG-Bench Hard (BBH). State-of-the-art models achieve near-perfect scores on many tasks in BBH, thus diminishing its utility. To address this limitation, we introduce BIG-Bench Extra Hard (BBEH), a new benchmark designed to push the boundaries of LLM reasoning evaluation. BBEH replaces each task in BBH with a novel task that probes a similar reasoning capability but exhibits significantly increased difficulty. 
    We evaluate various models on BBEH and observe a (harmonic) average accuracy of 9.8\% for the best general-purpose model and 44.8\% for the best reasoning-specialized model, indicating substantial room for improvement and highlighting the ongoing challenge of achieving robust general reasoning in LLMs. We release BBEH publicly at: \url{https://github.com/google-deepmind/bbeh}.
\end{abstract}

\maketitle

\section{Introduction}
\begin{wrapfigure}{H}{0.57\textwidth}
\vspace{-45pt}
\centering
\includegraphics[width=0.57\textwidth]{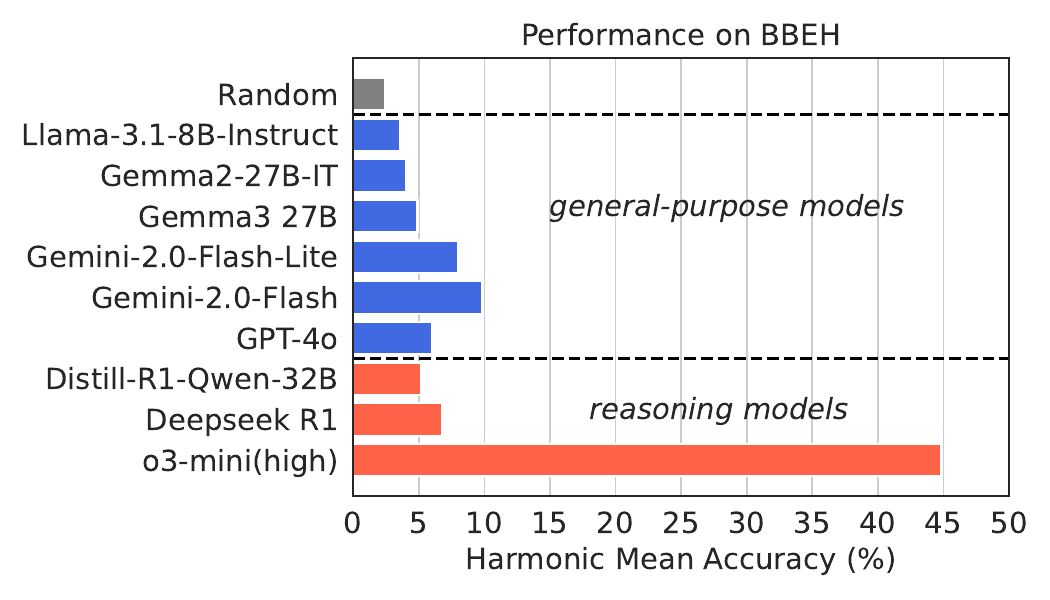}
\caption{Model performances on BBEH (harmonic mean over individual task performances).}
\label{fig:pull}
\vspace{-10pt}
\end{wrapfigure}

Recent research has made significant strides in evaluating the reasoning capabilities of large language models (LLMs), but the focus has been disproportionately skewed towards math/science and coding. This emphasis is likely driven by the availability of challenging benchmarks in these domains and the relative ease of evaluating quantitative solutions. However, reasoning encompasses a far broader spectrum of cognitive skills, including logical deduction, temporal and spatial understanding, commonsense reasoning, and even the ability to comprehend humor.\blfootnote{Llama and Qwen experiments in this paper were conducted only by parties outside of Google.}

To assess these diverse facets of reasoning, the community has relied on the BIG-Bench benchmark \citep{srivastava2022beyond}, specifically its more challenging subset, BIG-Bench Hard (BBH) \citep{suzgun2022challenging}. BBH has served as the de facto standard for evaluating general reasoning in LLMs due to its versatility and the wide array of reasoning skills it probes.  However, the rapid advancements in LLM development has led to a saturation of BBH, with state-of-the-art models achieving over 90\% accuracy. This performance ceiling renders BBH less effective in discriminating between the reasoning abilities of the latest generation of LLMs.

To address this challenge and push the boundaries of LLM evaluation, we introduce BIG-Bench Extra Hard (BBEH), a new benchmark designed to assess advanced reasoning capabilities. BBEH builds upon BBH by replacing each of the 23 tasks from BBH with a novel counterpart that probes similar reasoning capabilities, but exhibits significantly increased difficulty. Solving the tasks in BBEH requires even further reasoning skills than the problems in BBH. These skills include, but are not limited to, many-hop reasoning, learning on the fly, finding errors in reasoning traces, processing long-context inputs and finding (multi-)needles in a haystack, going against strong prior, dealing with long-range dependencies, dealing with distractors and inducing patterns from examples. By presenting LLMs with problems that demand a wider array of reasoning capabilities, BBEH aims to provide a more accurate measure of their general reasoning abilities.

We provide a comprehensive evaluation of several state-of-the-art LLMs on BBEH. Our results demonstrate that BBEH presents a significant challenge even for the most advanced models. Specifically, we observe a ceiling accuracy of 23.9\% for the best general-purpose model and 54.2\% for the model reasoning-specialized model, highlighting the substantial headroom for improvement in robust general reasoning for LLM. Furthermore, we conduct a detailed failure analysis, revealing intriguing failure modes for both general-purpose LLMs and models specifically designed for reasoning and thinking. These findings offer valuable insights into the current limitations of LLMs and provide guidance for future research aimed at enhancing their reasoning capabilities.

\section{Related Work}
There has been significant emphasis on using LLMs for mathematical and scientific reasoning. This has led to the popularity and proliferation of math- and STEM-based evaluations, such as \citep{hendrycks2020measuring, gsm8k}, and more recently, \citep{glazer2024frontiermathbenchmarkevaluatingadvanced, phan2025humanitysexam}. However, the generalizability of mathematical reasoning skills to broader domains remains unclear. Indeed, attempts to make existing benchmarks more robust—for example, \cite{mirzadeh2024gsm}—have highlighted an overall lack of robustness and logical reasoning capabilities. Several benchmarks have also been developed to address specific areas of reasoning, including temporal reasoning \citep{xiong2024large, beniwal2024remember, dhingra2022time}, spatial understanding \citep{bohnet2024exploringbenchmarkingplanningcapabilities, yamada2023evaluating, mirzaee2021spartqa, shi2022stepgame}, commonsense reasoning \citep{zellers2019hellaswag, talmor2018commonsenseqa, sakaguchi2021winogrande}, and logical reasoning \citep{saparov2022language, tafjord2020proofwriter, saparov2023testing,parmar2024logicbench}. However, these benchmarks tend to focus narrowly on specific domains, leading to potential evaluation biases if a more holistic view of model capabilities is not considered. To address this limitation, several benchmarks have been developed to integrate multiple tasks into a single evaluation framework, including \citep{wang2018glue, wang2019superglue, weston2015towards, lu2023mathvista, kazemi2024remi, hendrycks2020measuring,wang2024mmlu,parmar2024logicbench,srivastava2022beyond}. Our work builds on this line of research, introducing a new set of challenging tasks for future model evaluation and performance improvement. The multi-task nature of our benchmark with fine-grained tasks each focused on some reasoning skills enables model developers to discover and analyze failure modes in further depth.
Note that while private initiatives such as ChatBot Arena \citep{huggingfaceChatbotArena} and the SEAL leaderboard \citep{scaleSEALLeaderboards} conduct model evaluations across various aspects, they may suffer from several potential issues as pointed out in \cite{bansal2024peekingbehindclosed}. Our benchmark provides an open evaluation framework with an automatic and deterministic scoring mechanism, ensuring full transparency and reproducibility for the broader research community.

\section{Going Beyond BIG-Bench Hard} \label{sec:bbh}
BIG-Bench, and its later subset BIG-Bench Hard (BBH), have served the research community for several years as \emph{de facto} benchmarks for measuring general reasoning capabilities. Here, we mainly focus on BBH which has been favored in the recent years. We believe three key factors in the success of BBH were the following: 1- it was challenging even for the frontier models at the time, thus helping reveal model shortcomings and areas for improvement, 2- despite being composed of various tasks, all the questions in BBH were in a way that allowed for simple correctness verification of model responses with a few simple rules, and 3- perhaps most importantly, it tested for a vast array of reasoning skills making it a good proxy for measuring general reasoning. The latter factor is in contrast to many existing reasoning benchmarks that test only for a few of these skills as explained in the related work section.

Despite the great success of BBH and its widespread use, it also exhibited some key limitations. Firstly, a random chance baseline already has a high performance on BBH due to the high number of tasks with limited options (8/23 of the tasks have binary labels, and another 5/23 have at most 5 options to choose from). Secondly, some of the problems in BBH can be solved using shortcuts without solving the problem (for example, in the \emph{geometric shapes} task, whenever there are three \emph{L} commands the answer is \emph{triangle}). Thirdly, while real-world reasoning problems typically require processing large amounts of input texts, the input lengths of the problems in BBH are often quite short with a macro average of 700 characters across tasks. Fourthly, since the problems were made to be challenging for the LLMs of the time, they typically require only few hops of reasoning. And finally, while BBH tests for a quite large and diverse set of skills, the set can be greatly expanded to include even more reasoning skills.
The fast pace of improvement in the reasoning ability of the latest LLMs coupled with the limitations of the BBH dataset outlined above has led to BBH getting saturated with latest models achieving accuracies in the range of 90+ percent on it, thus causing BBH to disappear from the latest model evaluation reports.

We create a challenging benchmark for measuring the general reasoning capability of LLMs by using BBH as a guide: we preserve the positive aspects of BBH and lift the limitations as much as possible. Specifically, we create a benchmark that is challenging for the frontier models, the correctness of the model responses can be verified automatically despite being composed of several sub-tasks, the random chance baseline has a much lower success rate and the shortcuts are removed from the problems as much as possible, problems require processing longer inputs and require significantly more hops of reasoning to be solved, and, perhaps most importantly, covers a wide array of reasoning skills including those in the original BBH dataset and beyond. 
Through studying the 23 tasks in BIG-Bench Hard (BBH), we identify that for a reasoner to perform well on BBH it needs to have at least the following broad reasoning skills: 1- temporal understanding, 2- spatial and geometric understanding, 3- commonsense understanding, 4- humour understanding, 5- causal understanding, 6- reasoning about world entities and events, 7- deductive logical reasoning, 8- reasoning through linguistic knowledge, 9- counting and filtering, 10- data structures and algorithms, and 11- performing arithmetic operations.

\tikzset{
  header node/.style={%
    rectangle split,
    rectangle split parts=2,
    draw,
    rounded corners,
    thick,
    rectangle split part fill={#1, white},
    text width=8cm,
    align=center,
    font=\scriptsize,
    execute at begin node=\strut
  }
}

\tikzset{
  large header node/.style={%
    header node=#1, 
    text width=8cm,
    align=left,
    minimum height=8cm
  }
}

\begin{figure}
\begin{tikzpicture}[->, >=stealth, node distance=1.5cm and 2cm, xshift=-10cm]

\node[header node=chartreuse, yshift=1.3cm] (example1) {
    \textbf{Spatial Reasoning} 
    \nodepart[large header node]{second}
    You have been given a diamond tile map consisting of N rows [...] There is a unique object placed at each vertex. [...] You are initially at the top corner where you see a football. Then you move down-right for one step and see a shampoo. Then you move down-left for one step and you see a cat. [...] Then, you jump to a random vertex V where you see a bear. Then you move [...] Then you move up-left and you see a shampoo. Then you jump back to the random vertex V and do the following moves: down-left, down-left, down-right, up-left, down-left, up-left. What will you find?
  };
\node[header node=chartreuse, right=0.3cm of example1] (example2) {
    \textbf{Buggy Tables} 
    \nodepart[large header node, text height=10cm]{second}
    I have a table with 30 rows (including the header) and 18 columns. The table was converted to Markdown format as follows:
    \textbf{<TABLE IN MARKDOWN FORMAT>}. However, the code used to convert the table into Markdown format was buggy and mistakenly replaced some values with \texttt{"ERROR"}. The correct values for those cells in row-order are respectively as follows: $[9, 10, \text{null}, 12, \dots, 29]$.
    Compute the absolute difference between the mean of \texttt{coding\_minutes} and \texttt{exercise\_minutes}, considering only the days where: 1- The number of meetings was greater than 2.
  };
  
\node[header node=chartreuse, below=0.3cm of example1] (example3) {
    \textbf{Causal Understanding} 
    \nodepart[large header node]{second}
     Reagent X is being added to a tank. At each time interval, a drop of Reagent X is being added. At one point the tank overflows. Is adding the last drop of Reagent X a necessary cause for the tank to overflow?
    
  };
\node[header node=chartreuse, right=0.3cm of example3] (example4) {
    \textbf{Word Sorting} 
    \nodepart[large header node]{second}
    Consider a new alphabet whose letters have the same order as the English alphabet, except that r and p are swapped. Sort the following words with the new alphabet and separate them with comma: syndrome, therefrom, [...], specifications.
  };
  
\node[header node=chartreuse, below=0.3cm of example3] (example5) {
    \textbf{Multistep Arithmetic} 
    \nodepart[large header node]{second}
    % \Mstrut{\rule[-7pt]{0pt}{15pt}}
    Consider the following new operations:\\
    \quad\quad\quad\quad\quad\quad\quad $a \, [] \, b = [...]$
    \[
    a \, @ \, b = 
    \begin{cases} 
        (a - b) * b, & \text{if } a\, []\, b < 2 \\ 
        (b - a) * a, & \text{otherwise}
    \end{cases}
    \]
    
    For brevity, we use \( a \, \langle op1 \rangle \langle op2 \rangle \, b \) to denote \( (a \, op1 \, b) \, op2 \, b \). \newline
    Let A = ((((1 @*+ 4) <>+[] (-4 *<>* -1)) @; ((-1 <> six) ; (2 ;* one))) @@@ (((five ;- five) []@@ (-8 - one)) ; ((two +; -5) +[]- (three - -8)))). Let B = [...]. Let C = [...]. Compute: A + B - C.
  };

\node[header node=chartreuse, right=0.3cm of example5] (example6) {
    \textbf{BoardgameQA} 
    \nodepart[large header node]{second}
    A few players are playing a boardgame. The current state of the game is as follows. The bee has a football with a radius of 15 inches. The chihuahua has a smoothie. [...]\newline
    Rule1: Anyone who [...] has to pay \(\$\$\$\) to the beaver. \newline
    Rule2: If the stork [...], then it does not pay \(\$\$\$\) to beaver. \newline
Rule 2 is preferred over Rule 1. If a rule is preferred over the other, it means whenever both of them can be applied to derive new conclusions and those conclusions contradict with each other, we should go with the conclusion from the rule with higher preference.
    
What is the truth value of the statement: "Does the finch shout at the mermaid?"
  };
  
  \node[header node=chartreuse, below=0.3cm of example5] (example7) {
    \textbf{Dyck Language} 
    \nodepart[large header node]{second}
    You are given an initial Dyck language sequence and the steps, provided as thoughts, that were used to arrive at the closing bracket sequence in the Dyck language.  
    Your job is to identify the first step that was a mistake.
    
    Task: Complete the rest of the sequence, making sure that the parentheses are closed properly.
    
    Input: 
    ( \, < \, < \, > \, [ \, (
    
    Thoughts: \textbf{Thought 1:} We should process each input one by one and keep track of the stack configuration. \textbf{Thought 2:} stack: empty \textbf{Thought 3:} \( ( \); stack: \( ( \) [\dots]
  };
 
 \node[header node=chartreuse, right=0.3cm of example7] (example8) {
    \textbf{Time Arithmetic} 
    \nodepart[large header node]{second}
    Let the answer to Q1 be \(X\) and the answer to Q2 be Y.

    Q1: Clara and William were born on \texttt{2015-Aug-24} and \texttt{2016-May-20} respectively.  
    When William was 326 days old, how old was Clara in days?
    
    Q2:  <TEXT OF THE QUESTION>
    
    Define: X' = X + 3, \quad Y' = Y - 568 \newline
    Q3:   
    Alan and Mary tried a new restaurant on \texttt{Aug Y', 1997} and really liked it.  
    They decided to go to the same restaurant every \(X'\) days.  
    If today is \texttt{Oct 11, 1997}, when is the next time they will go to that restaurant?
  };

  \node[header node=chartreuse, below=0.3cm of example7] (example9) {
    \textbf{Shuffled Objects} 
    \nodepart[large header node]{second}
  Alice, Bob, Claire, Dave, Eve, Fred, and Gertrude are dancers at a square dance. At the start of a song, they each have a partner: Alice is dancing with Rodrigo, Bob is dancing with Jamie, Claire is dancing with Patrick, Dave is dancing with Lola, Eve is dancing with Izzi, Fred is dancing with Melissa, and Gertrude is dancing with Karl. Throughout the song, the dancers often trade partners. First, Fred and Bob switch partners. Then, Fred and Claire switch partners (let's call it Action 1). Then, Claire and Bob switch partners (let's call it Action 2). [...] Then, Action 1 repeats. Then, Fred and Claire switch partners. [...] At the end, who is Fred dancing with?
  };
 
 \node[header node=chartreuse, right=0.3cm of example9] (example10) {
    \textbf{Geometric Shapes} 
    \nodepart[large header node]{second}
    Suppose we draw this SVG path element: M -18.79152,1.82096 L -8.48528,-8.48528 M -32.29803,24.74825 [...] . Out of the following shapes: 
    \begin{enumerate}
        \item rectangle that is not a square and with no diagonals drawn
        \item square with no diagonals drawn
        \item [...]
    \end{enumerate}
    Which one(s) can be viewed when the lines in the SVG are visualized? Options: (A) 1 and 3 (B) only 5 [...].
  };
\end{tikzpicture}
\caption{Sample Questions from 10 tasks. The text has been shortened due to lack of space. \label{fig:samples}}
\end{figure}

For our benchmark, we wish to preserve the high diversity of the reasoning skills and capabilities from BBH and also expand upon it by including more of the following skills. 1- \textbf{Many-hop reasoning:} The ability to solve a problem that require many steps/hops, 2- \textbf{Very long-range dependency:} the ability to remember and use information that have been provided or concluded much earlier in the context, 3- \textbf{Going against strong prior:} the ability to reason through a problem even when it does not match the prior beliefs (\cite{mckenzie2023inverse} report inverse scaling behaviours in such cases), 4- \textbf{Learning on the fly:} the able to learn something new on the fly (i.e. from the information in the context) and apply it to solve a problem, 5- \textbf{Dealing with distractors:} The ability to identify the parts of information provided that are necessary for solving the problem, and not getting distracted by the redundant information, 6- \textbf{Long-context:} the ability to reason through a long input context and stitching different pieces of input together, 7- \textbf{Needle in a haystack:} the ability to find small pieces of relevant information from a large input (e.g., finding one value from a large table), 8- \textbf{Finding errors in reasoning traces:} the ability to identify the errors in a chain of reasoning not produced by the model itself, 9- \textbf{Inductive reasoning: } the ability to induce patterns from a number of examples and being able to apply the same pattern to a new instance of a problem, 10- \textbf{Contraint satisfaction:} the ability to understand constraints and find a solution that satisfies them, 11- \textbf{Compositional understanding:} the ability to solve multiple independent problems that are fused into one problem (\cite{hosseini2024not} and \cite{miner2024scheherazade} report higher failure rates than expected in the compositional case), and 12- \textbf{Knowledge-intense reasoning:} the ability to reason in domains where a great amount of domain knowledge is needed.

\section{BIG-Bench Extra Hard}

We create BIG-Bench Extra Hard (BBEH), a dataset that tests the general reasoning capability of models on a wide array of reasoning skills. To this end, we build on the success of BBH and replace each of the 23 tasks in BBH with another task that is in a similar reasoning domain and tests for similar (or more) skills, but is more challenging compared to the original one. Replacing each task with another one in the same domain that tests for similar capabilities ensures that we preserve the high diversity of original BBH dataset. In Table~\ref{table:tasks}, we outline a high-level description of the new tasks in BBEH, how they have been constructed and which task from BBH the replace, and what reasoning skills they target. The benchmark contains $200$ questions per task, except for the \emph{Disambiguation QA} task where we have $120$ questions. For more details about the tasks and some intuitions from the experimental results and model failure modes, see Appendix~\ref{sec:details}. Samples from a few of our tasks are provided in Figure~\ref{fig:samples}.

\begin{table}[th!]
\centering
\small
\resizebox{\textwidth}{!}{  
\begin{tabular}{p{2.5cm}|p{18cm}|p{7cm}}
\toprule
\textbf{BBEH task} & \textbf{Summary of changes made and the task it replaces from BBH} & \textbf{Main reasoning skills} \\
\midrule
Boardgame QA & Based on \cite{kazemi2023boardgameqa} but with larger reasoning depth. Requires many hops of deductive logical reasoning, and also requires the model to learn a specific type of conflict resolution on the fly. Replaces Logical Deduction from BBH which needed only a few simple steps of deductive logic. & Deductive reasoning, learning on the fly, many-hop reasoning.  \\
\midrule
Boolean Expressions & Requires determining the truth value of an expression whose operands could themselves be textual/mathematical sub-expressions that evaluate to True or False. Replaces the \emph{Boolean Expressions} task from BBH which can be easily solved through one line of python code. & Logical reasoning, many-hop reasoning. \\ 
\midrule
Buggy Tables & Requires understanding and reconstructing a large buggy table given the description of the bug, and then computing some conditional queries on it. Replaces \emph{Penguins in a table} from BBH which required simple operations over small, clean tables. & Data structures, learning on the fly, needle in haystack \\ 
\midrule
Causal Understanding & A subset of the causal stories in \cite{nie2023moca} and improved examples from \cite{kiciman2023causal}. One subtask is focused on testing causal judgment and the other on testing the ability to reason about necessary and sufficient causes.  Replaces the \emph{Causal Judgement} task from BBH. &  Causal judgement/reasoning, logical reasoning, counterfactual reasoning\\ 
\midrule
Disambiguation QA & A task created by the authors, requiring pronoun disambiguation over longer and more challenging text compared to the original \emph{Disambiguation QA} task in BBH. & Commonsense understanding, linguistics knowledge. \\ 
\midrule
Dyck Language & Involves finding errors in (potentially) faulty solutions to closing a sequence of brackets. It comes from \cite{tyen2023llms} and replaces the \emph{Dyck languages} task from BBH which requires properly closing brackets as opposed to finding errors.  & Data structures, finding errors in reasoning traces. \\ 
\midrule
Geometric Shapes & Requires identifying the shapes drawn by a series of SVG commands. Each set of commands can draw multiple shapes and may involve many distracting commands that do not participate in any shape. Replaces the \emph{Geometric Shapes} from BBH which involved identifying a single shape. & Spatial reasoning, geometric understanding, dealing with distractors.  \\ 
\midrule
Hyperbaton & Requires inducing correct adjective order given examples on a new variant of English, and properly apply it to new examples. Replaces \emph{Hyperbaton} from BBH which required simply knowing the correct adjective order in English. & Inductive reasoning, going against strong prior, linguistic knowledge.  \\
\midrule
Linguini & Comes from \cite{sanchez2024linguini} and requires linguistic reasoning and inductive reasoning to learn about a new language given some examples and then properly apply those learnings. Replaces \emph{Salient Translation Errors} from BBH which involved simpler linguistic understanding.  & Inductive reasoning, linguistic knowledge. \\ 
\midrule
Movie Recommendation & Given a number of sets of movies, the task is to determine which set has movies that are all likely to be liked by a specific group of people. Replaces \emph{Movie Recommendation} from BBH which required simple next movie recommendation. & Reasoning through knowledge \\ 
\midrule
Multi-step Arithmetic & Requires learning new arithmetic operations and their compositions on the fly, and apply them to evaluate long expressions. Replaces the \emph{Multi-step Arithmetic} task from BBH which involved simple arithmetic over basic operations. & Learning on the fly, many-hop reasoning. \\ 
\midrule
New Yorker Cartoon Caption (NYCC) & Comes from \cite{hessel2022androids, zhang2024humoraimassivescale}  and requires selecting the funniest caption for an image. We adopt the variant that predicts the best caption only given the textual description of the image. This replaces the \emph{Ruin Names} task from BBH which involved simpler humour understanding. & Humour understanding, commonsense understanding. \\ 
\midrule
Object Counting & Requires counting the number of objects of a certain type given a very long list of various objects and in presence of many types of distractors. Replaces the \emph{Object Counting} from BBH which required simple counting in a short context. & Long-context, (multi-)needle in a haystack, dealing with distractors. \\ 
\midrule
Object Properties & Requires keeping track of a large collection of objects with various properties while they go through multiple rounds of modification. Replaces \emph{Colored Objects} from BBH which required only recognizing the color of some objects. & Temporal track keeping, long-range dependency.  \\ 
\midrule
SARC Triples & Requires understanding sarcasm in Reddit posts and replies. Each problem requires determining the sarcastic-ness of three post/reply pairs. Replaces the \emph{Snark} task in BBH which required simpler sarcasm understanding. & Commonsense understanding, sarcasm understanding, simple compositional reasoning. \\ 
\midrule
Shuffled Objects & A long-context variant of the original \emph{Shuffled Objects} from BBH which may also require remembering very long-range information. & Temporal track keeping, long-context, (multi)-needle in a haystack, long-range dependency. \\ 
\midrule
Spatial Reasoning & Adopts the SpatialLLMEval \citep{yamada2023evaluating} dataset which requires spatial reasoning over various complex patterns. We expanded the dataset with problems that require many-hops of reasoning and require both forward reasoning from the premises to the goal and backward thinking from the goal to the premises. Replaces the \emph{navigation} task from BBH which requires much simpler spatial understanding of navigation signals. & Spatial understanding, many-hop reasoning, long-range dependency.  \\ 
\midrule
SportQA & Comes from \cite{xia2024sportqa} and requires reasoning combined with a high amount of sports knowledge. We use the hardest subset that contains compositional questions, where each problem contains one main question and multiple sub-questions. Replaces \emph{Sport Understanding} from BBH which needed much simpler reasoning over sport knowledge. & Knowledge-intensive reasoning, compositional reasoning. \\ 
\midrule
Temporal Sequences & Requires finding proper meeting times given multiple calendars (each corresponding to a temporal sequence) and various constraints. Replaces the \emph{Temporal Sequence} task from BBH which involves understanding only a single sequence. & Temporal understanding, constraints satisfaction. \\ 
\midrule
Time Arithmetic & Comes from the Test of Time benchmark \citep{fatemi2024test} and involves various operations over various representations of date/time. We created a compositional version of this task following \citep{hosseini2024not}. Replaces the \emph{Date Understanding} task from BBH which involved significantly simpler operations over dates. & Temporal reasoning, compositional understanding. \\ 
\midrule
Web of Lies & Requires many-hop reasoning to predict the truthfulness of a set of people, and contains two subsets: one coming from the variant used in LiveBench \citep{white2024livebench} and one novel variant that involves cases where the truthfulness of some individuals remains unknown but new conclusions can be drawn from it nevertheless. Replaces the \emph{Web of Lies} from BBH which involved simpler cases of this problem. & Logical reasoning, many-hop reasoning. \\ 
\midrule
Word Sorting & Contains two subtasks: 1- sorting over a modified alphabet order, which goes against the strong prior of the model, and 2- finding errors in sorting traces. Replaces the original Word Sorting task which required simple sorting. & Apply algorithms, Going against strong prior, Finding errors in reasoning traces.  \\
\midrule
Zebra Puzzles & Puzzles that require various logical deductions to be solved. We add distracting clues to the puzzles to make them more challenging. The dataset is an expanded version of the one from \cite{shah2024causal} and replaces \emph{Formal Fallacies} from BBH which requires understanding logic and formal fallacies in much simpler setups. & Constraint satisfaction, many-hop reasoning, dealing with distractors, long-range dependency.  \\ 
\bottomrule
\end{tabular}
}
\caption{The tasks in BBEH in alphabetical order of the names, a high-level description of what they test for, the reasoning capabilities that they probe, and the task from BBH that they replace.}
\label{table:tasks}
\end{table}

A key challenge in creating benchmarks is ensuring they remain difficult for frontier models. This is particularly true for reasoning benchmarks, given the rapid progress in the field over the past year, and especially for BBEH, which comprises 23 distinct tasks, each requiring careful design. To ensure our tasks challenge frontier models, we adopted a semi-adversarial approach. We selected two strong reference models: one general-purpose and one specialized in reasoning. We iteratively increased task difficulty while keeping in mind the extra skills that we wanted our benchmark to test for, evaluating the reference models on each new iteration. If a task proved insufficiently challenging, we either replaced it with another task or added extra types of difficulty and re-evaluated until the difficulty level was satisfactory. We used Gemini 1.5 Flash \citep{team2024gemini} as our general-purpose reference model and the Gemini Thinking Experimental model as our reasoning-specialized reference model (initially the December 2024 version but later changed to the January 2025 version, known as Gemini-2.0-Flash-Thinking-Exp-01-21). These models were chosen for their performance and the speed of generating outputs, which facilitated rapid iteration during task construction. We iterated on each task until both reference models achieved an accuracy below 70\%.

In most cases, we tried to use the reference models only as a black box that provided feedback on the difficulty of our tasks. In some cases, however, making tasks more difficult required looking into the approach adopted by the model. As an example, the original "Boolean Expression" task in BBH required models to evaluate the truth value of expressions such as \emph{(not True) or False}. Our initial attempt to increase difficulty involved creating longer expressions with significantly more clauses. However, our reference model achieved high accuracy regardless of the number of clauses. While initially this seemed surprising, upon investigating the model's approach, we discovered it cleverly used Python to solve the problem by directly evaluating the expression: \emph{result = <expression>; print(result)}. Thus, adding more clauses did not have much effect in increasing difficulty. Our next step was to prevent the model from using Python. We achieved this by replacing some "True" and "False" clauses with sentences that evaluated to the same truth value (e.g., replacing "True" with "The capital of Canada is Ottawa.").

Given the similarity of the high-level approach in creating LLM reasoners (architecture, training phases, etc.), we believe our semi-adversarial benchmark construction can lead to a benchmark that is also challenging for non-reference models. This is confirmed by the experimental results in the following sections. However, this approach also has some notable limitations. Firstly, the choice of the reference model will unavoidably bias the benchmark towards certain types of failure modes. For instance, had our reference model not used code to solve the multi-hop Boolean expressions, we might have stopped there, resulting in a task too easy for models that appropriately trigger code. We tried to mitigate this as much as possible by using strong reference models, and by avoiding over-engineering to the reference model failures. Secondly, since the benchmark is created adversarially with respect to the reference models, a fair comparison of the reference and non-reference models may not be possible. We expect this limitation to be temporary and be resolved when newer versions of the reference models become available.

\textbf{BBEH Mini:} Besides reporting results on BBEH, we also report results on a smaller subset called \emph{BBEH Mini} which contains $460$ examples overall ($20$ examples randomly selected from each task). This subset can be used for faster and cheaper experimentations.

\begin{figure}[t]
  \centering
  \includegraphics[width=\columnwidth]{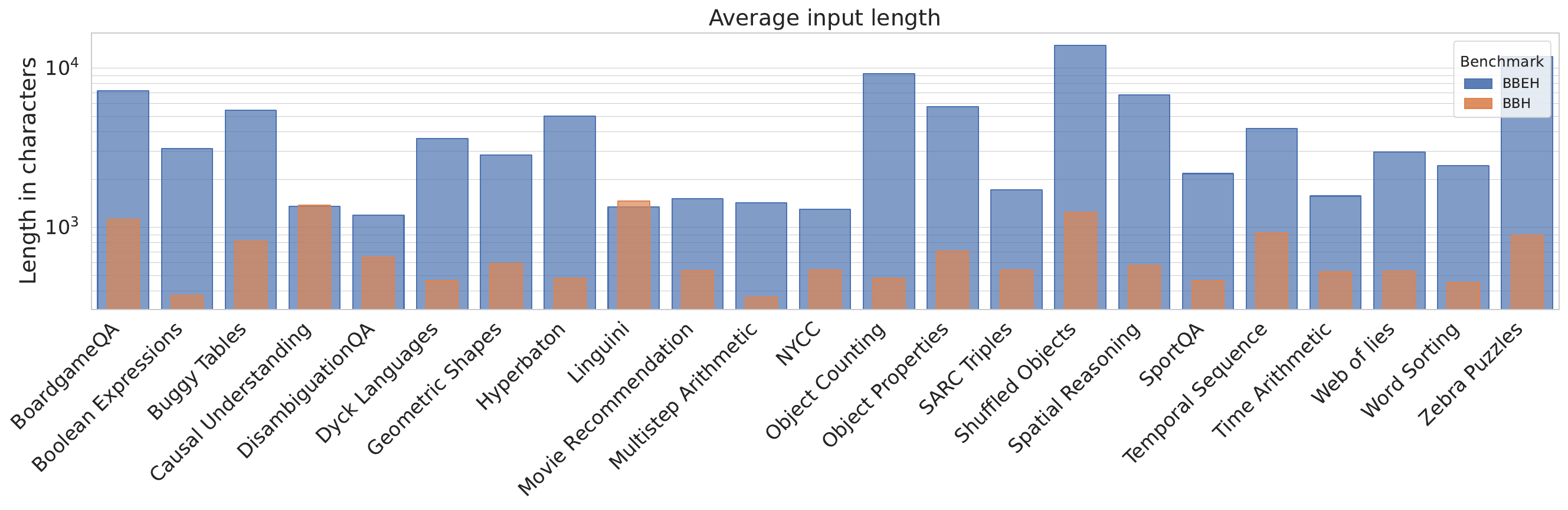} %

  \caption{%
  \label{fig:input_length} %
  A comparison of the average input lengths of the tasks in BBEH to their counterparts in BBH.
  }
\end{figure}

\section{Results and Analyses}
We start by analyzing BBEH and comparing it against its counterpart, BBH. We then report results on BBEH for various models and compare their performances. Then, we provide some extra analysis of the results revealing interesting insights about where reasoning-specialized and larger models gain more and where they gain less compared to general-purpose and smaller models respectively. We also provide a large body of observations and insights from task-specific results in Appendix~\ref{sec:details}.

\subsection{BBEH Analysis}
\textbf{Input/Context Length:} As mentioned in Section~\ref{sec:bbh}, the problems in the original BBH dataset are mostly short. On the contrary, the problems in BBEH tend to be quite long and require a great amount of input processing by the models. Figure~\ref{fig:input_length} compares the average input lengths of each of the tasks in BBEH with their counterpart from BBH. From the figure, one can observe how input lengths have increased for almost all the tasks (except two), sometimes quite significantly. The macro average context length of the tasks in BBEH is about six times bigger than that of BBH.

\begin{figure}[t]
  \centering
  \includegraphics[width=\columnwidth]{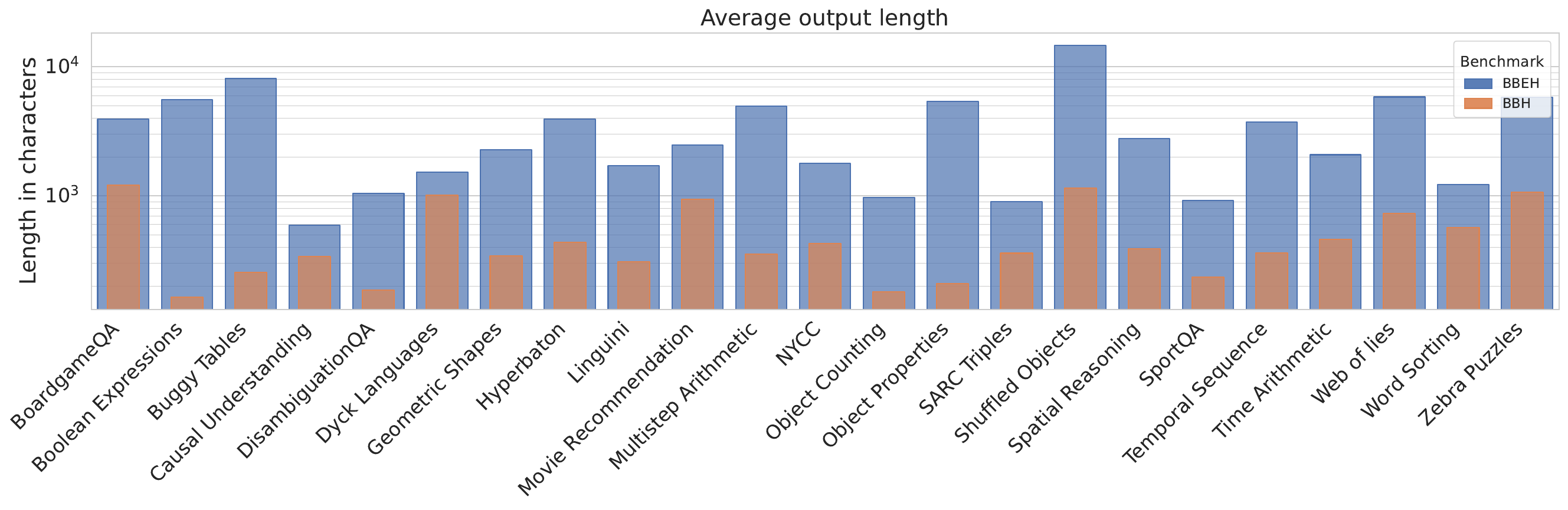} %

  \caption{%
  \label{fig:output_length} %
  A comparison of the average output lengths of the Gemini 2.0 Flash responses for each of the tasks in BBEH and their counterparts in BBH.
  }
\end{figure}

\textbf{Required Amount of Thinking:} Many of the problems in BBH only require few hops of reasoning, sometimes not requiring a great amount of thinking. As a proxy for measuring the amount of thinking required by BBEH and compare it to BBH, we compare the average length of the outputs generated by a fixed model (Gemini 2.0 Flash) for the two datasets. The results are presented in Figure~\ref{fig:output_length}. From the figure, we can observe that the average length of the output has significantly increased for every single one of the tasks in BBEH compared to their counterpart in BBH, thus providing evidence that the problems in BBEH may require much more thinking. The macro average output length of the responses for tasks in BBEH is about seven times bigger than that of BBH.

\begin{table*}[th!]
\centering
\resizebox{\textwidth}{!}{  
\begin{tabular}{c|cccccccccc|ccc}
\toprule
\textbf{Tasks/Models} & \rotatebox{90}{\textbf{Random}} & \rotatebox{90}{\textbf{Qwen-2.5-7B-Instruct}} & \rotatebox{90}{\textbf{Llama 3.1 8b Instruct}} & \rotatebox{90}{\textbf{Gemma2 27b IT}} &
\rotatebox{90}{\textbf{Gemma3 4b}} &
\rotatebox{90}{\textbf{Gemma3 12b}} & \rotatebox{90}{\textbf{Gemma3 27b}} & \rotatebox{90}{\textbf{Gemini 2.0 Flash-Lite}} & \rotatebox{90}{\textbf{Gemini 2.0 Flash}} & \rotatebox{90}{\textbf{GPT4o}} & \rotatebox{90}{\textbf{Distill R1 Qwen 32b}} & \rotatebox{90}{\textbf{DeepSeek R1}} & \rotatebox{90}{\textbf{o3-mini (high)}} \\
\toprule
BoardgameQA & 33.3 & 31.0 & 31.5 & 39.5 & 34.5 & 33.0 & 37.0 & 29.5 & 42.5 & 41.0 & 36.0 & 75.5 & 53.0 \\
Boolean Expressions & 20.0 & 22.0 & 18.0 & 25.0 & 23.0 & 25.5 & 22.0 & 24.0 & 27.0 & 22.5 & 17.5 & 55.5 & 67.0 \\
Buggy Tables & 0.0 & 1.5 & 0.0 & 0.5 & 0.5 & 0.0 & 0.0 & 1.5 & 3.5 & 0.5 & 0.5 & 4.5 & 59.5 \\
Causal Understanding & 38.0 & 40.0 & 37.0 & 45.5 & 46.0 & 49.0 & 51.5 & 52.5 & 52.0 & 54.0 & 54.5 & 54.5 & 54.0 \\
DisambiguationQA & 21.0 & 34.2 & 36.7 & 45.0 & 33.3 & 41.7 & 49.2 & 50.0 & 48.3 & 51.7 & 52.5 & 50.0 & 58.3 \\
Dyck Languages & 1.4 & 1.0 & 4.5 & 2.0 & 3.5 & 8.5 & 4.5 & 6.5 & 14.0 & 8.0 & 18.0 & 56.0 & 55.0 \\
Geometric Shapes & 6.2 & 41.5 & 25.5 & 31.0 & 18.0 & 32.5 & 26.5 & 30.0 & 35.0 & 22.5 & 4.5 & 1.5 & 52.5 \\
Hyperbaton & 0.0 & 0.5 & 2.0 & 4.0 & 2.0 & 2.5 & 3.5 & 6.5 & 4.5 & 7.5 & 3.0 & 6.0 & 32.0 \\
Linguini & 0.0 & 2.5 & 3.0 & 7.0 & 1.0 & 6.0 & 9.0 & 12.5 & 15.5 & 15.5 & 6.0 & 19.5 & 17.0 \\
Movie Recommendation & 10.0 & 24.5 & 30.0 & 40.0 & 35.0 & 44.5 & 55.0 & 51.5 & 59.5 & 61.0 & 46.0 & 59.5 & 84.0 \\
Multistep Arithmetic & 0.0 & 0.0 & 0.5 & 0.0 & 0.0 & 0.5 & 1.5 & 7.5 & 9.5 & 5.5 & 36.0 & 46.5 & 73.0 \\
NYCC & 10.0 & 13.0 & 13.0 & 13.5 & 7.0 & 11.5 & 15.0 & 13.5 & 11.0 & 23.0 & 10.5 & 20.0 & 16.0 \\
Object Counting & 0.0 & 0.0 & 0.0 & 0.0 & 0.0 & 0.5 & 0.0 & 4.0 & 11.0 & 6.5 & 4.0 & 76.5 & 90.0 \\
Object Properties & 1.6 & 0.0 & 0.5 & 0.0 & 0.5 & 1.5 & 0.5 & 0.5 & 1.5 & 0.0 & 0.0 & 0.0 & 56.5 \\
SARC Triples & 12.5 & 17.5 & 16.5 & 21.0 & 14.0 & 26.0 & 24.0 & 27.0 & 37.5 & 38.5 & 22.0 & 28.5 & 24.0 \\
Shuffled Objects & 14.3 & 8.0 & 9.5 & 12.0 & 1.0 & 7.0 & 5.0 & 15.0 & 9.0 & 14.0 & 2.0 & 6.0 & 49.5 \\
Spatial Reasoning & 5.2 & 0.0 & 1.0 & 7.0 & 4.0 & 10.5 & 13.0 & 10.5 & 18.5 & 14.0 & 14.5 & 37.0 & 48.5 \\
SportQA & 0.0 & 5.0 & 1.5 & 10.0 & 2.5 & 12.5 & 20.0 & 18.5 & 23.0 & 25.0 & 19.5 & 29.0 & 26.5 \\
Temporal Sequences & 0.0 & 0.5 & 9.5 & 1.5 & 1.0 & 0.0 & 1.5 & 1.0 & 0.5 & 0.0 & 0.5 & 0.0 & 68.5 \\
Time Arithmetic & 0.5 & 18.5 & 4.0 & 15.5 & 9.5 & 23.5 & 46.5 & 45.0 & 48.0 & 45.5 & 56.5 & 77.0 & 76.5 \\
Web of Lies & 5.5 & 2.5 & 5.5 & 6.5 & 12.0 & 22.0 & 21.0 & 14.0 & 18.5 & 14.5 & 13.0 & 29.5 & 43.0 \\
Word Sorting & 4.3 & 4.0 & 2.5 & 3.5 & 4.5 & 6.0 & 7.5 & 12.5 & 26.0 & 22.0 & 36.0 & 68.0 & 77.5 \\
Zebra Puzzles & 15.4 & 19.0 & 2.5 & 23.0 & 10.0 & 19.5 & 30.0 & 32.0 & 44.5 & 32.0 & 1.5 & 8.0 & 67.5 \\
\midrule
BBEH & 2.4 & 3.0 & 3.6 & 4.0 & 3.4 & 4.5 & 4.9 & 8.0 & 9.8 & 6.0 & 5.2 & 6.8 & 44.8 \\
\bottomrule
\end{tabular}
}
\caption{The performance of various models on the individual tasks and overall on BBEH.}
\label{tab:main-results}
\end{table*}

\subsection{Model Evaluations}
\textbf{Models:} We evaluate various models on BBEH and compare their performance across individual tasks and on the entire dataset. Specifically, we experiment with models from the following families: Llama 3.1 \citep{dubey2024llama}, Qwen 2.5 \cite{yang2024qwen2}, Gemma2 \citep{team2024gemma}, Gemma3 \citep{team2025gemma}, Gemini 2.0, GPT4o (the latest version, 2024-11-20, at the time of the experiments) \citep{achiam2023gpt}, DeepSeek R1 and the Distilled model in Qwen 32b \citep{guo2025deepseek}, and o3-mini (high)\footnote{\url{https://openai.com/index/openai-o3-mini/}}.

\textbf{Metric:} Given the highly versatile use-cases of the current LLM reasoners, they should be capable across the board to excel at real-world problems and be robust general reasoners. However, we find that micro and macro averages (which are often used for benchmarks composed of multiple tasks), fail to capture this crucial aspect. These metrics are susceptible to distortion by outlier performance, potentially presenting a misleadingly optimistic assessment when a model excels in a limited subset of tasks while faltering in others. To address this, we employ the (adjusted) harmonic mean\footnote{To deal with zero values, we add a value of $1$ to all accuracy numbers.} as our primary evaluation metric for BBEH. The harmonic mean provides a more conservative and balanced representation of overall performance, effectively penalizing models with significant performance disparities across different tasks, thereby aligning more closely with the requirement for consistent, general reasoning capabilities. For BBEH Mini, due to the small number of examples per task, we adopt micro average.

The results for each task and on the entire dataset for each model is presented in Table~\ref{tab:main-results} for BBEH and in Table~\ref{tab:mini-results} for BBEH Mini. According to the results, we make several interesting observations. Firstly, we observe a large headroom not only for the individual tasks, but also for BBEH overall. The best performance for the general-purpose models is at 9.8\% harmonic mean accuracy. The reasoning-specialized models are expectedly performing better than the general-purpose models on the benchmark, but the best performance for these models is still at 44.8\% on BBEH. Despite the adversarial construction, the reference Thinking model achieves a harmonic mean accuracy of 20.2\% on BBEH. Note that while we calibrated the difficulty with respect to two reference models so their accuracies fall below 70\%, the difficulty mostly carries to other models too with o3-mini (high) exceeding 70\% accuracy only on 4 out of 23 tasks, DeepSeek R1 exceeding it only on 3 out of 23 tasks, and other models never exceeding it. Note that some model accuracies are even below random performance. Upon checking, we observe that these are mostly cases where models could not solve the problem in their effective output token lengths and started degenerating after a point, so no final answer could be extracted from their solution.

Secondly, for completeness, we also report micro average accuracies in Table~\ref{tab:micro-average} (in Appendix) according to which we observe that the best general-purpose model has a micro average accuracy of 23.9\% and the best reasoning-specialized model has an accuracy of 54.2\%. Interestingly, while DeepSeek R1 performs better than all general-purpose models in terms of micro average accuracy, given its low performance on some of our tasks it performs worse than two of the general-purpose models in terms of harmonic mean accuracy.

Thirdly, as mentioned in Section~\ref{sec:bbh}, the problems in the original BBH dataset suffered from having a small output space, thus allowing for a random baseline to have a high performance. In Table~\ref{tab:main-results}, we provide the results of a random baseline for each of the tasks in BBEH and the entire dataset. As can be viewed, the random baseline has a performance of $8.4\%$ for BBEH which leaves substantial room for comparing models of various size.

Finally, looking at the accuracies of the models on various tasks, we can see that various models are good at different types of reasoning. For example, DeepSeek R1 significantly outperforms other models on BoardgameQA, o3-mini (high) significantly outperforms other models on Temporal Sequences and Object Properties, GPT4o significantly outperforms other models on NYCC, and GPT4o and Gemini 2.0 Flash significantly outperform other models on SARC Triples.

\begin{table}[t]
\centering
\begin{tabular}{c|cccccccc|ccc}
\toprule
\footnotesize
\textbf{Tasks/Models} & \rotatebox{90}{\textbf{Llama 3.1 8b Instruct}} & \rotatebox{90}{\textbf{Gemma2 27b IT}} & \rotatebox{90}{\textbf{Gemma3 4b}} & \rotatebox{90}{\textbf{Gemma3 12b}} & \rotatebox{90}{\textbf{Gemma3 27b}} & \rotatebox{90}{\textbf{Gemini 2.0 Flash-Lite}} & \rotatebox{90}{\textbf{Gemini 2.0 Flash}} & \rotatebox{90}{\textbf{GPT4o}} & \rotatebox{90}{\textbf{Distill R1 Qwen 32b}} & \rotatebox{90}{\textbf{DeepSeek R1}} & \rotatebox{90}{\textbf{o3-mini (high)}}\\
\toprule
\makecell{BBEH Mini\\(Micro Average)} & 11.5 & 15.0 & 13.3 & 14.3 & 17.4 & 22.2 & 27.0 & 23.5 & 15.4 & 37.2 & 56.7 \\
\bottomrule
\end{tabular}
\caption{The performance of various models on BBEH Mini (micro average).}
\label{tab:mini-results}
\end{table}

\subsection{Further Analyses of the Results}
\begin{figure}[t]
  \centering
  \includegraphics[width=\columnwidth]{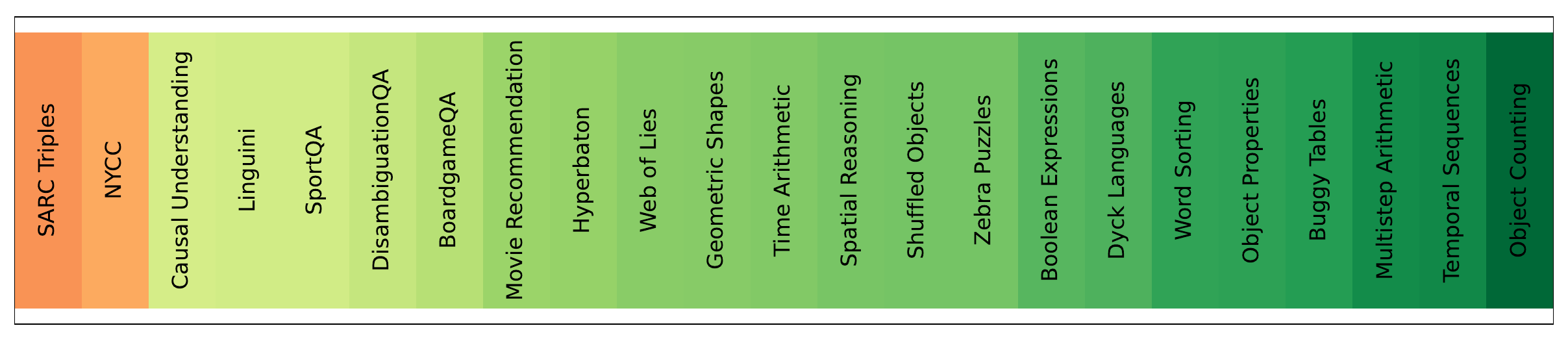} %

  \caption{%
  \label{fig:heatmap} %
  \textbf{Performance gains (absolute) of o3-mini (high) over GPT-4o on BBEH tasks.} Tasks are ordered by the magnitude of improvement, with green signifying substantial gains and yellow/red signifying minimal or negative gains. }
  
\end{figure}

\textbf{General-Purpose vs Reasoning Models:} With the introduction of reasoning models that leverage test-time compute for thinking, a tremendous jump in performance was observed on reasoning tasks involving math and coding. For example, on the AIME2024 dataset, the performance of GPT4o was 13.4\%, but the o1 model increased it to 83.3\% and o3-mini (high) increased it further to 87.3\%. Here, we examine whether the same is true for various types of general reasoning. In Figure~\ref{fig:heatmap}, we compare o3-mini (high) and GPT4o, as examples of reasoning and general models respectively, on each of the tasks from BBEH and sort the tasks ascending based on how much o3-mini (high) gains over GPT4o. We observe that the tasks that gain the most are those involving counting, planning, arithmetic, and data structures and algorithms. Whereas the tasks that gain the least (or sometimes negatively) are mostly those involving commensense, humour, sarcasm, and causation. Our results indicate that reasoning models achieve the most significant gains when applied to formal problems and demonstrate limited progress in handling the softer reasoning skills which are typically needed for complex, real-world scenarios.

\begin{figure}[t]
  \centering
  \includegraphics[width=\columnwidth]{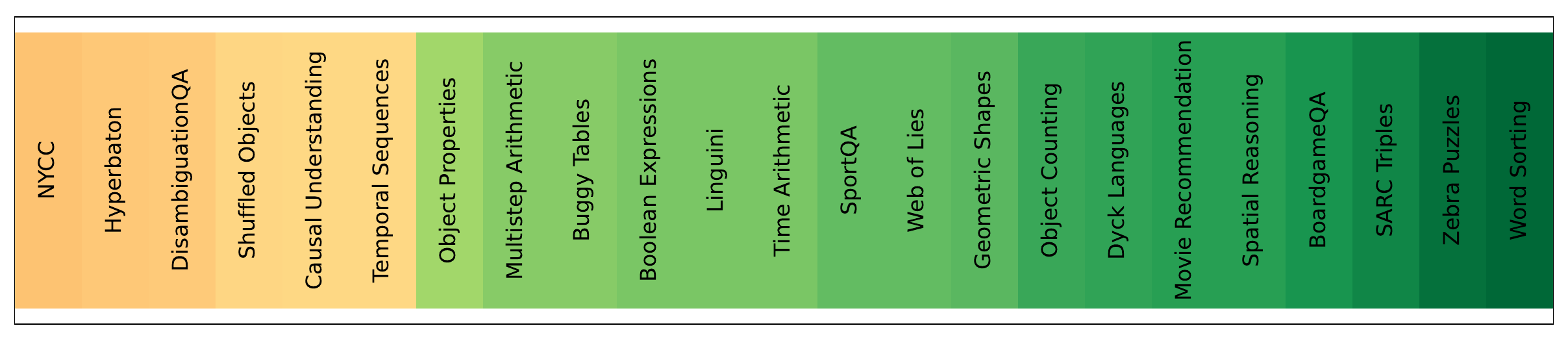} %

  \caption{%
  \label{fig:size_heatmap} %
  \textbf{Performance gains (absolute) of Gemini 2.0 Flash over Gemini 2.0 Flash-Lite on BBEH tasks.} Tasks are ordered by the magnitude of improvement, with green signifying substantial gains and yellow/red signifying minimal or negative gains. }
\end{figure}

\textbf{The Effect of Model Size:} We also explore the effect of model size on the performance of the models across various types of general reasoning. In Figure~\ref{fig:size_heatmap}, we compare Gemini 2.0 Flash against Gemini 2.0 Flash-Lite on different tasks from BBEH and sort the tasks ascending based on how much Flash gains over Flash-Lite\footnote{Since in some cases these models perform below random chance due to not generating an extractable final answer, to reduce noise we take the maximum of their performance and the random chance performance for each task.}. While the signal is not as clear as the case where we compared general models against reasoning models, we still observe that the tasks related to humour, commonsense, and causal reasoning are the ones with the least gains, and tasks requiring many-hop reasoning or applying algorithms are the ones with the largest gains. A particular exception is the SARC Triples task which is a sarcasm understanding and where the gains are large. This could in part be due to the fact that each example in SARC Triples is a composition of three sub-questions, and larger models may be better at dealing with such composite questions.

\begin{figure*}[t]
    \centering
    \begin{subfigure}[ht]{0.45\textwidth}
        \centering
        \includegraphics[width=\textwidth]{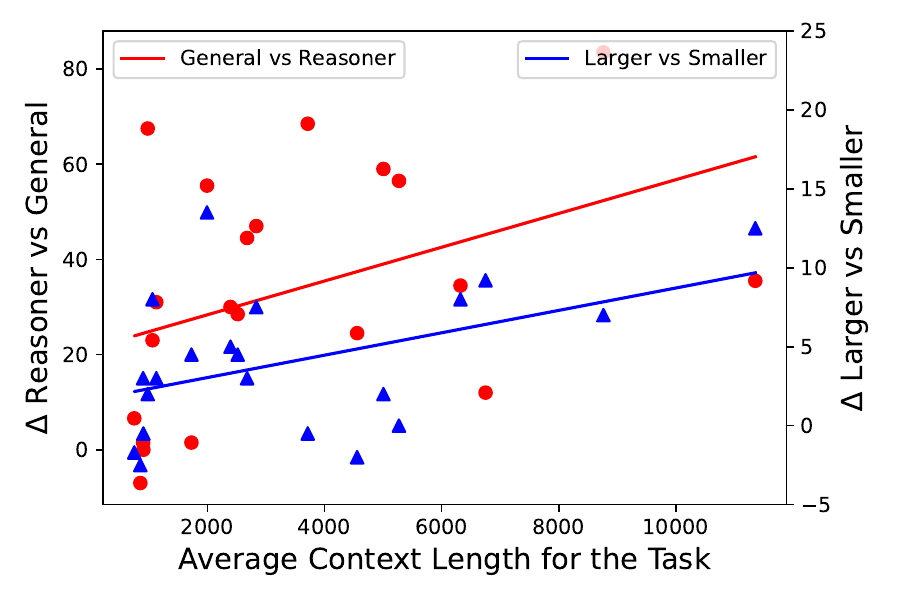}
    \end{subfigure}%
    ~
    \begin{subfigure}[ht]{0.45\textwidth}
        \centering
        \includegraphics[width=\textwidth]{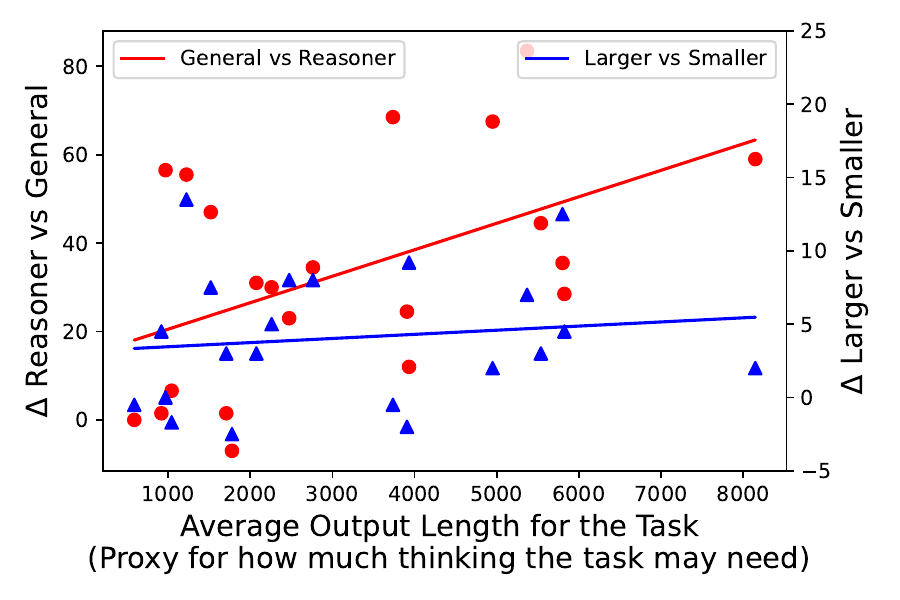}
    \end{subfigure}
    \caption{\textbf{Performance gains as a function of (left) context length and (right) output length (proxy for required thinking).} A scatter plot and trendlines for the gains obtained by a reasoning-specialized model (o3-mini high) vs a general-purpose model (GPT4o) and a larger model (Gemini 2.0 Flash) vs a smaller model (Gemini 2.0 Flash-Lite), as a function of (left) the average context lengths and (right) the average output lengths (same values as in Figure~\ref{fig:output_length}), for the tasks in BBEH.}
    \label{fig:context_output_vs_perf}
\end{figure*}

\textbf{The Effect of Context Length and Required Thinking:} The tasks in BBEH come at different average context lengths (see Figure~\ref{fig:input_length}) and may require different amount of thinking (as shown using the output length proxy in Figure~\ref{fig:output_length}). We use this property to understand the effect of context length and required thinking on reasoning vs general models, and on larger vs smaller models. To this end, in Figure~\ref{fig:context_output_vs_perf} we compare the performance of o3-mini (high) vs GPT4o and Gemini 2.0 Flash vs Gemini 2.0 Flash-Lite as a function of average context lengths of the tasks and average output length as a proxy for required thinking\footnote{We removed the \emph{Shuffled Objects} task for this analysis as some models ran out of effective tokens and started degenerating, and this was adding noise to the analysis.}. We observe that the gains of o3-mini tend to increase compared to GPT4o both when context length increases and when the required thinking increases, showing how reasoning models may have improved across both directions compared to general models. For Gemini 2.0 Flash vs Gemini 2.0 Flash-Lite, we see a similar increase in gains when the context length increases, but the curve for the case of increased thinking remains mostly flat.

\section{Conclusion}
Recent advances in LLM reasoning has made these models reach near ceiling performance on existing general reasoning benchmarks such as BIG-Bench and its harder variant BBH, and shifted focus toward other types of more focused reasoning. However, substantial distance remains before we can claim these models posses true mastery of diverse reasoning skills. To rekindle the pursuit of truly robust and versatile LLM reasoners, we presented BIG-Bench Extra Hard (BBEH), a significantly more challenging successor to BBH. This new benchmark, meticulously crafted to amplify the difficulty of existing tasks while preserving their core diversity, reveals a stark reality: even the most advanced LLMs still grapple with fundamental aspects of general reasoning. BBEH provides a crucial stepping stone, reigniting the challenge and offering a more rigorous platform for future research aimed at unlocking the full potential of LLMs in complex, real-world applications.

\section*{Acknowledgements}
We acknowledge the help from Vahab Mirrokni, Tania Bedrax-Weiss, Don Metzler, Javad Hosseini, Phoebe Kirk, and Katherine Tong.

\bibliography{MyBib}

\pagebreak

\appendix

\begin{table}[th!]
\centering
\resizebox{\textwidth}{!}{ 
\begin{tabular}{c|cccccccccc|ccc}
\toprule
\footnotesize
\textbf{Models} & \rotatebox{90}{\textbf{Random}} & \rotatebox{90}{\textbf{Qwen-2.5-7B-Instruct}} & \rotatebox{90}{\textbf{Llama 3.1 8b Instruct}} & \rotatebox{90}{\textbf{Gemma2 27b IT}} & \rotatebox{90}{\textbf{Gemma3 4b}} & \rotatebox{90}{\textbf{Gemma3 12b}} & \rotatebox{90}{\textbf{Gemma3 27b}} & \rotatebox{90}{\textbf{Gemini 2.0 Flash-Lite}} & \rotatebox{90}{\textbf{Gemini 2.0 Flash}} & \rotatebox{90}{\textbf{GPT4o}} & \rotatebox{90}{\textbf{Distill R1 Qwen 32b}} & \rotatebox{90}{\textbf{DeepSeek R1}} & \rotatebox{90}{\textbf{o3-mini (high)}} \\
\toprule
\makecell{BBEH\\(Micro Average)} & 8.4 & 12.5 & 10.6 & 14.8 & 11.0 & 16.3 & 18.8 & 19.7 & 23.9 & 22.3 & 19.2 & 34.9 & 54.2 \\
\bottomrule
\end{tabular}
}
\caption{The micro average performance of various models on BBEH.}
\label{tab:micro-average}
\end{table}

\section{Detailed Description of the Tasks and Task-Specific Insights from Experiments } \label{sec:details}
Here, we describe in detail how each of the 23 new tasks in BBEH have been created. Moreover, we provide interesting task-specific insights from our experiments.

\begin{table}[t]
\centering
\footnotesize
\begin{tabular}{c|cccccc}
\toprule
\textbf{Models} & Gemma2 27b IT & Gemini 2.0 Flash-Lite & Gemini 2.0 Flash & GPT4o & DeepSeek R1 & o3-mini (high) \\
\midrule
\textbf{Unknown \%} & 77.6 & 67.4 & 73.3 & 82.4 & 39.7 & 65.5 \\
\bottomrule
\end{tabular}
\caption{Percentage of \emph{unknown} predictions on BoardgameQA for different models (only one third of the labels are \emph{unknown}).}
\label{tab:boardgame-qa-unknowns}
\end{table}

\subsection{BoardgameQA}

BoardgameQA \citep{kazemi2023boardgameqa} is a benchmark where given a defeasible theory (a set of input facts, possibly contradictory rules, and preferences over the rules), and a question about that theory, the task is to do multi-hop reasoning and conflict resolution over the input theory to answer the question. The final answer to the question is either `proved` (if the statement in the question derives from the theory), `disproved` (if the negation of the statement in the question derives from the theory), or `unknown` (if neither the statement in the questions nor its negation derives from the theory). With three labels per question, a random baseline has an accuracy of ~33.3\%. Conflicts may arise when two rules such as $R1: a \rightarrow c$ and $R2: b \rightarrow \neg c$ are both activated leading to different beliefs about the truth value of the variable $c$. However, preferences over the rules is provided in the input question and in the case of conflicts, the derivation from the rule with the higher preference must be concluded (e.g., if $R1$ is preferred over $R2$ and they both apply, then we conclude $c$ is true).

One of the parameters controlling the difficulty of the problems in this benchmark is the depth, corresponding to the number of hops of reasoning that must be done to compute the truth value of the statement in the question. We use the code from the paper and generate tasks with depths 6, 7 and 8. We made some changes in the prompt to clarify the task for the model so that it works in zero-shot setting. That includes: \emph{Answer 'proved' if it can be proved, 'disproved' if it can be disproved, and 'unknown' if it can neither be proved nor disproved} as well as \emph{A rule is only applicable if all of its antecedents can be proved.}. We then uniformly sampled prompts across depths and labels to created the final set.

Through analyzing the model outputs, we observe that for this task, models tend to over-predict that the truth value of a statement is \emph{unknown}. The percentage of \emph{unknown} predictions for our models (when at least one of the three labels was predicted) is presented in Table~\ref{tab:boardgame-qa-unknowns}. Note that only one-third of the problems have an \emph{unknown} label. An \emph{unknown} label is typically predicted when the model cannot find a way to either prove or disprove the statement from the facts and rules. Therefore, we observe a failure mode for the state-of-the-art models: they struggle to search the space of facts and rules and find a proof, despite there being one. While similar observations have been previously made about models without chain-of-thought \citep{saparov2024transformers}, our results apply to the case with chain-of-thought.

\subsection{Boolean Expressions} 
This task requires determining the truth value of a statement that is composed of logical operands such as $True$ and $False$ as well as other textual or mathematical statements that evaluate to True or False. To create this task, we first randomly create expressions containing only True and False operands and three logical operators: \emph{and}, \emph{or}, and \emph{not}. We create this in a bottom-up fashion where we generate smaller sub-expressions and then combine them with logical operators. Once a large enough expression is created, we replace some of the $True$ and $False$ operands with statements that evaluate to True or False. These could be mathematical expressions such as \texttt{24 - 2 is greater than 48 / 2} (which evaluates to False) or textual statements such as \texttt{The capital of Canada is Ottawa} (which evaluates to True). In both cases, we select these statements from a predefined set. While determining the truth value of each of these statements in isolation may be easy for many models, including these statements makes it more difficult for models; otherwise, they can simply solve the problem by generating a single line of python code. 

We generate five expressions using the approach outlined above, four of which evaluate to False and one of which evaluate to True. The job of the model is then to find the expression that evaluates to True. Since this is a five-way question, the random chance accuracy is 20\%.

\subsection{Buggy Tables}
The objective in this task is to be able to respond to conditional queries over tabular data, where the information in the table are presented in a buggy way but the description for the bug is also presented so that the model can reconstruct the original table based on that. As an example, we provide a row-major/column-major format of the table where the null values have been mistakenly removed, but we also provide the positions of the null values in the original table so one can reconstruct the table given the two pieces of information. As another example, we provide a buggy version of the table where some random values are appended at the end of each row or each column, but we also specify how they have been added so one can use this information to remove them and reconstruct the original table. As yet another example, we provide a markdown format of the table that mixes each two rows of the table into one row, but also provide an explanation of how each two rows have been merged into one so that the original table can be reconstructed based on that information. Examples of conditional queries include computing some statistics (count, sum, mean, stdev, median) of some columns while only considering rows where some columns have some specific values. 

\begin{table}[th!]
\centering
\small
\resizebox{\textwidth}{!}{  
\begin{tabular}{l|p{10cm}|l|p{6cm}}
\toprule
\textbf{Ex. ID} & \textbf{Input} & \textbf{Output} & \textbf{Gemini 2.0 Flash Response}  \\
\midrule
36 & Question: Suzy and Billy are working on a project that is very important for our nation's security. The boss tells them both: ``Be sure that you are here at exactly 9 am. It is absolutely essential that you arrive at that time.'' Both Billy and Suzy arrive at 9 am. As it happens, there was a motion detector installed in the room where they arrived. The motion detector was set up to be triggered if at least one person appeared in the room at the same time. So the motion detector went off. Did Billy cause the motion detector to go off? Reply Yes or No based on the answer the majority of people would give. If you think people would be split roughly 50-50 between Yes and No then reply Ambiguous. 
& 
Ambiguous
&
The question asks whether Billy caused the motion detector to go off. The motion detector was triggered if at least one person was present. Both Billy and Suzy were present at 9 am. Suzy's presence also triggered the motion detector. Since the motion detector would have gone off regardless of whether Billy was there, it cannot be definitively said that Billy caused the motion detector to go off. The answer is: Ambiguous.  \\
\midrule
123 & Question: A group of students who lived on the same floor of a dormitory obtained a copy of the final exam for their biology class. The students cheated on the test. One student, John Granger, did not go along with the group. Granger deviates from the norm and refuses to cheat on the test. The biology class comprises 80 students and is graded on a curve such that 20 people will receive a grade of A, 20 a grade of B, 20 a grade of C, and 20 students will receive a D. Granger\'s score was the 20th-highest score in the class, which means he was the last student to receive a grade of A. The 21st student was a pre-med student who received a B and, as a result, missed the GPA cutoff she needed to get into the medical school she was hoping for by .07 GPA points. Did Granger cause the student to fail to meet the medical school cutoff? Reply Yes or No based on the answer the majority of people would give. If you think people would be split roughly 50-50 between Yes and No then reply Ambiguous. 
& 
No
&
Here's a step-by-step analysis: 1. John's Score: John earned the 20th highest score, placing him at the very top of the ``A" grade range. 2. The Pre-Med Student's Score: The pre-med student scored the 21st highest score, placing her at the top of the ``B" grade range. 3. Impact on the Curve: If John had cheated and scored higher, he could have potentially been in the ``A" range with a higher score. This would have potentially shifted the scores on the curve down. 4. The Student's Failure: It is possible, though not guaranteed, that if John had cheated and scored higher, the curve could have shifted down. 5. Conclusion: The pre-med student failing the cutoff is related to the grading curve. If a number of students cheated, it changed the overall distribution. It is highly probable that, without cheating, the pre-med student would have gotten the higher grade. The answer is: Yes.  \\
\midrule
136 
&
Question: Alice and Zoe work for the same company. They work in different rooms and both of them sometimes need to access the central computer of the company. Unbeknownst to everybody, if two people are logged in to the central computer at the same time, some spam emails containing dangerous viruses are immediately deleted from the central computer. In order to make sure that one person is always available to answer incoming phone calls, the company issued the following official policy: Alice is the only one permitted to log in to the central computer in the mornings, whereas Zoe is the only one permitted to log in to the central computer in the afternoons. One day, violating the official policy, Zoe logs in to the central computer at 9 am. The same day, Alice also logs in at 9 am. Immediately, some spam emails containing dangerous viruses are deleted from the central computer. Did Alice cause some spam emails containing dangerous viruses to be removed from the central computer? Reply Yes or No based on the answer the majority of people would give. If you think people would be split roughly 50-50 between Yes and No then reply Ambiguous.  
&
Yes 
&
The question asks if Alice caused the deletion of spam emails. The scenario states that if Alice and Zoe are logged in simultaneously, the spam emails are deleted. Both Alice and Zoe logged in at the same time. Thus, Alice's action, combined with Zoe's, resulted in the deletion of the spam emails. The answer is: Yes" \\
\bottomrule
\end{tabular}
}
\caption{Causal understanding: examples of causal judgment queries and correct or incorrect reasoning traces. Example ID corresponds to the index of the question in the causal understanding sub-benchmark.}
\label{table:failures_causal_judgment}
\end{table}

\subsection{Causal Understanding}
In BBEH we replace the original causal judgement task in BBH with a set of questions that assess both (i) \emph{causal judgement} (142 queries) and (ii) the ability to determine \emph{necessary and sufficient causes} (58 queries). In this section we describe how these different sets of questions are obtained.

\paragraph{Causal Judgement}
These queries are based on the 144 causal stories included in the MoCa benchmark \citep{nie2023moca}, which partially overlap with the sets of questions originally included in BBH. In MoCa, short stories obtained from cognitive science papers were given to 25 human annotators who had to \textit{judge} whether, based on the given story, a certain person or event caused a certain outcome. The task was phrased as a binary task with Yes/No answers, and the ground truth label was assigned according to the label chosen by the majority of humans. 

However, the stories included complex normative and logical factors, and for many of them there was a large degree of disagreement among the human annotators. In cases where the human raters strongly disagreed on the answer (defined as having a difference of at most 20\% between the ``Yes'' and ``No'' answers among the annotators), questions were additionally tagged as ``Ambiguous''. Based on this, we constructed the renewed task to have 3 possible labels: Yes, No and Ambiguous. The label ambiguous was assigned to the 46 questions originally tagged as ``Ambiguous'' in MoCa. 
For instance, the label for example 36 in Table \ref{table:failures_causal_judgment} was changed from Yes to Ambiguous, as 15 human annotators replied Yes while 10 replied No. In this example we have that, on the one hand Billy was asked to be in the room at 9am and cannot be given the fault of entering the room and triggering the alarm. On the other hand, the alarm was set to be triggered if at least one person appeared in the room, thus both Billy and Suzy could be assigned the responsibility for the detector going off. Both explanation could be considered valid. For the remaining questions, the label was kept unchanged to Yes/No. 

With the above re-definition of ground truth labels, models were then asked to correctly identify the way humans, as a group, would answer the question, thus testing alignment with human causal intuitions. To reflect this, we added the following instructions to each query: 

\begin{tcolorbox}[colback=black!5!white,colframe=black!75!black,title=Prompt Instructions]
Reply Yes or No based on the answer the majority of people would give. \\
If you think people would be split roughly 50-50 between Yes and No then reply Ambiguous.
\end{tcolorbox}

Finally, to ensure consistency in terms of number of queries included for the other tasks, we removed 2 stories included in the original set (these correspond to question 17 and 19 at \url{https://moca-llm.github.io/causal_stories/1/}). The final set includes 45 Ambiguous questions and 48 and 49 questions with Yes and No labels respectively.

\paragraph{Necessary and Sufficient Causes} We complemented the causal judgement stories with 58 examples testing reasoning about necessary and sufficient causes given a description of a set of events (example scenario). These examples were obtained by modifying those in \cite{kiciman2023causal} to correct ambiguities and incorrect outputs. 

The first 30 examples in \cite{kiciman2023causal} were constructed from 15 scenarios introduced in different studies over the years to critique actual causality \citep{halpern2016actual} definitions from the literature \citep{kueffner2021comprehensive}. Each scenario is associated with a question about necessary cause and a question about sufficient cause, specifically “Is \{Actor\} a necessary cause of \{Event\}?” and “Is \{Actor\} a sufficient cause of \{Event\}?”. To test LLM memorization issues, \cite{kiciman2023causal} constructed 28 additional examples from 14 new scenarios obtained by adapting the original ones to a chemistry lab setting involving reagents, mixtures and crystals.  

Three experts of causal reasoning analysed each example and suggested minimal changes to resolve ambiguities, typos, and incorrect outputs. The main ambiguities that were identified relates to the use of \{Actor\}, which was substituted with the associated action. Each example was discussed to reach an agreement on the changes, this resulted in changing the outputs of six examples (see Table \ref{table:vignettes}).

\paragraph{Model Outputs Analysis}
Analysing the performance that Gemini 2.0 Flash achieves on the causal understanding task reveals that this model answers correctly to 45\% of the causal judgment queries (random performance is 33\%) and 71\% of the queries about necessary and sufficient causes (random performance is 50\%). 

Focusing on the causal judgment queries, most of the errors are in questions for which the ground truth label is Ambiguous (44 mistakes out of 45 examples) or No (24 mistakes out of 49 examples), with only 10 mistakes out of 48 examples for the Yes label. This reveals the difficulty the model has in determining an absence of causal relationships, and in dealing with ambiguities and the existence of different possible answers. This behaviour has been previously observed across other causal reasoning benchmarks (see e.g. \cite{romanou2023crab}) where models have been found to infer stronger causal relationships than those that humans perceive. Another interesting failure mode is that the model sometimes exhibits a lack of understanding of the \emph{normative} aspect of causal judgements as made by humans: humans tend to ascribe causality more easily when the causal factor is unusual in a statistical sense, or when it violates an established rule or behavioral norm \citep{halpern2008defaults,phillips2015unifying,kominsky2015causal,halpern2015graded,icard2017normality}. For instance, in example 136 of Table \ref{table:failures_causal_judgment}, the event can only occur if both Alice and Zoe log on simultaneously to a computer -- which is indeed what happened. However, Alice was allowed to log on while Zoe violated an established rule when she logged in. In such situations, humans tend to say that Alice did not cause the event, while Zoe did. The model fails to capture this nuance and instead it applies straightforward causal reasoning and concludes that Alice is a cause.

Looking at the reasoning traces for Gemini 2.0 Flash responses to questions on sufficient and necessary causes reveals that, while the model achieves very close performances in terms of precision and recall, it fails on identifying sufficient causes (11 errors out of 28 examples) more often than identifying necessary causes (6 errors out of 30 examples). Interestingly, the model correctly recalls the definition of necessary and sufficient causes in most of the responses and uses counterfactual reasoning to consider alternative scenarios. Despite this, the model often fails at interpreting some of the causal links described in the scenarios (in example 155 of Table \ref{table:vignettes} the model interprets the input as implying that ``flowers would likely die in hot weather whether the neighbor waters or not'') or draws incorrect conclusions despite correct reasoning traces.

\begin{table}[th!]
\centering
\small
\resizebox{\textwidth}{!}{  
\begin{tabular}{l|p{8cm}|l|p{8cm}|l}
\toprule
\textbf{Example ID} & \textbf{Original Input} & \textbf{Original Output} & \textbf{ Modified Input} &  \textbf{Modified Output} \\
\midrule
149
& Two two-state switches are wired to an electrode. The switches are controlled by A and B respectively, and the electrode is attached to C. A has the first option to flip her switch. B has the second option to flip her switch. The electrode is activated and shocks C if both switches are in the same position. B wants to shock C, and so flips her switch iff A does. C gets an electric shock. Is A's action to flip the switch a necessary cause for C getting shocked? & No  & Two two-state switches are wired to an electrode. The switches are controlled by A and B respectively, and the electrode is attached to C. A has the first option to flip her switch. B has the second option to flip her switch. The electrode is activated and shocks C if both switches are flipped. B wants to shock C, and so flips her switch if and only if A does. C gets an electric shock. Is A's action to flip the switch a necessary cause for C getting shocked? & Yes \\ 
\midrule
154
& There are a left and a right window. Alice and Bob both order Carol to fire at the left window. Carol fires at the left window, shattering it. Commands from Alice always trump commands form Bob (e.g. if Bob would have ordered to fire at right window, Carol would still have fired at the left one.). Without a command Carol would not have fired at all. Is Alice a necessary cause for window shattering? & Yes & There are a left and a right window. Alice and Bob both order Carol to fire at the left window. Carol fires at the left window, shattering it. Commands from Alice always trump commands from Bob (e.g. if Bob would have ordered to fire at the right window, Carol would still have fired at the left one). Without a command Carol would not have fired at all. Is Alice ordering Carol to fire a necessary cause for the window shattering? & No\\
\midrule
155
& If there is hot weather, flowers will die. Watering prevents the flowers to die in hot weather. The neighbor does not water the flowers in her yard. The flowers die. Is neighbor's inaction a necessary cause for flowers' death? & No & If there is hot weather, flowers will die. Watering prevents the flowers from dying in hot weather. The neighbor does not water the flowers in her yard, the weather is hot and the flowers die. Is the neighbor's inaction a necessary cause for the flowers' death? & Yes\\
\midrule
170
&  If there is hot weather, flowers will die. Watering prevents the flowers to die in hot weather. The neighbor does not water the flowers in her yard. The flowers die. Is neighbor's inaction a sufficient cause for flowers' death? & Yes &  If there is hot weather, flowers will die. Watering prevents the flowers from dying in hot weather. The neighbor does not water the flowers in her yard, the weather is hot and the flowers die. Is the neighbor's inaction a sufficient cause for the flowers' death? & No\\
\midrule
177
& Reagent X is added to a beaker containing a  crystal. If Reagent X touches the crystal, the crystal dissolves. If Reagent X does not touch the crystal, Sam adds Reagent Y which leads to the crystal dissolving. Is Reagent X a necessary cause for crystal dissolving? & No & Reagent X is added to a beaker containing a crystal. If Reagent X touches the crystal, the crystal dissolves. If, when added, Reagent X does not touch the crystal, Sam adds Reagent Y, which leads the crystal to dissolve. Is adding Reagent X to the beaker a necessary cause for the crystal to dissolve? & Yes\\
\midrule
183
& There is a test tube on the left and a test tube on the right. Sam and Riya both order Frank to break the left test tube. Carol throws the left test tube, breaking it. Commands from Sam always trump commands form Riya (e.g. if Riya would have ordered to break the right test tube, Frank would still have thrown the left one.). Without a command Frank would not have acted at all. Is Sam a necessary cause for test tube breaking? & Yes & There is a test tube on the left and a test tube on the right. Sam and Riya both order Frank to break the left test tube. Frank throws the left test tube, breaking it. Commands from Sam always trump commands from Riya (e.g. if Riya would have ordered to break the right test tube, Frank would still have thrown the left one). Without a command Frank would not have acted at all. Is Sam's order a necessary cause for the test tube to break? & No\\
\bottomrule
\end{tabular}
}
\caption{Causal understanding: queries for which the output was changed with respect to the original dataset in \cite{kiciman2023causal}. Example ID corresponds to the index of the question in the causal understanding sub-benchmark.}
\label{table:vignettes}
\end{table}

\subsection{Disambiguation QA}
This task introduces a more challenging adaptation of the original DisambiguationQA task in BBH. The objective is to accurately determine the referents of ambiguous pronouns in complex sentences, or to explicitly identify instances of unresolvable ambiguity by responding 'ambiguous'. To enhance the task's difficulty and complexity, we constructed a dataset of 120 novel examples that are longer than those in BBH, require more referent disambiguation, and each question contains more options so the random chance performance is lower. These examples were constructed either by creating entirely new sentences or combining existing BBH instances. Ten annotators (all of them the authors of the paper) were tasked with creating these examples, each comprising a potentially ambiguous sentence, a single correct resolution statement, and several distractor options for a multiple-choice format. To ensure data quality, each example underwent a two-stage verification process. First, a separate annotator independently evaluated the correctness of the resolution. Discrepancies were then resolved through a third-party adjudicator or collaborative refinement by all three annotators. In cases where consensus could not be reached, the annotators jointly revised the example to achieve clarity and accuracy. This rigorous process resulted in 25 examples requiring modification. An example of an ambiguous sentence is provided below.

\begin{tcolorbox}[colback=black!5!white,colframe=black!75!black,title=Ambiguous Example]
Here is a sentence with pronoun(s) whose antecedent(s) can either be derived from the context or is ambiguous.\\
Sentence: While walking through the forest, John saw a deer and its fawn. It was beautiful.\\
Which of the following options correctly explains the antecedent(s) of the pronoun(s)?\\
(A) The deer was beautiful.\\
(B) The fawn was beautiful.\\
(C) The walk through the forest was beautiful.\\
(D) Ambiguous.
\end{tcolorbox}

\subsection{Dyck Languages}
This task comes from the BIG-Bench Mistake dataset \citep{tyen-etal-2024-llms}. It involves finding the first mistake in an existing chain-of-thought sequence, used to answer a Dyck Languages question in the original BBH dataset. In each example, the target answer is either the number where the first mistake occurred, or that there are no mistakes in the CoT sequence.
These CoT sequences are generated by prompting PaLM 2 Unicorn \citep{anil2023palm2technicalreport} on the original BBH dataset at temperature = 0. The prompts can be found at \url{https://github.com/suzgunmirac/BIG-Bench-Hard/blob/main/cot-prompts/dyck_languages.txt}. The newline is used as a stop token so that each intermediate step can be prepended with `Thought 1: ', `Thought 2: ', etc. Further information on the prompting and generation process can be found in \citet{tyen-etal-2024-llms}.

In the cases where there is an error in the trace and the model makes a mistake in identifying the first error, the mistake can occur due to two different reasons: 1- mis-classifying a correct reasoning step as erroneous before any error has occurred, and 2- missing the first error and identifying some later erroneous step. We looked into the breakdown of what percentage of the errors belong to each category when there is an error to be found and the model also identifies one of the steps as erroneous.

We find that the majority of the errors belong to the second category. Specifically, for o3-mini (high) 98.7\% of the errors belong to the second class, for Gemini 2.0 Flash all the errors belong to the second class, for Gemini 2.0 Flash-Lite 94.9\% and for GPT4o 96.8\% of the errors belong to the second class. This highlights a failure mode for the frontier models in that they can identify the correct reasoning steps, but fail to identify the ones that have errors.

\subsection{Geometric Shapes}
SVG is a language for drawing shapes. We use two basic commands: 1- $M (x, y)$ corresponding to moving to the $(x, y)$ coordinate, and 2- $L (x, y)$ corresponding to drawing a line from the current location to $(x, y)$. We use the shape outlines from GeomVerse \citep{kazemi2023geomverse}, a dataset of geometry questions involving multiple shapes that share some elements, which are specified as TikZ commands and convert them to SVG. We then ask the model to identify what shapes will be drawn if we visualize the SVG.

We consider two extra axes for difficulty: 1- we randomly break some lines segments into multiple colinear line segments, and 2- we add some extra lines such that they intersect at some points and those intersections form some shapes (in other cases, shapes are created using the full line segments and not at their intersection points). We then create four subsets for the task corresponding to the cross product of few vs many line breaks and intersect vs no intersect.

For o3-mini (high), which is the best performing model on this task, we observe that the accuracy for the \emph{few breaks} subset is 58\% while the accuracy for the \emph{many breaks} subset is 47\%; we also observe that the accuracy for the \emph{no intersect} subset is 72\% while for the \emph{intersect} subset is 33\%. This shows that both axes are adding to the difficulty of the task.

\subsection{Hyperbaton}

In English, a prescribed order governs the sequence of multiple adjectives preceding a noun. This order, generally, is: opinion → size → age → shape → color → origin → material → purpose. The BBH suite includes Hyperbaton, an adjective ordering task designed to evaluate a model’s linguistic knowledge—specifically, its understanding of adjective categories and adherence to the correct adjective ordering in English. In this task, models must choose the sentence with the correct adjective sequence from a pair of English sentences. Current general-purpose models excel at this task, reaching nearly 95\% accuracy (see Table~\ref{tab:comp-results} for Gemini 2.0 Flash performance on Hyperbaton task from BBH).

We replace Hyperbaton with a novel task that assesses inductive reasoning in addition to linguistic knowledge. This new task involves inducing the correct adjective order in a new variant of English, given a set of example sentences. Specifically, for each test sample, we create a unique English variant by randomly shuffling the standard adjective order and generating 50-250 sentences. Each sentence contains 1 to 3 adjectives preceding a noun. Models must then infer the correct adjective order for this variant and identify all correct sentences from 10 provided options. A “None of the above” option is included as a distractor. We observe that this variant already poses a challenge for latest general-purpose models. To further increase the difficulty, we ensure that the provided sentences only reveal partial adjective orderings, such as opinion, size, shape, color, purpose, age, and material, origin, without specifying the relative order of all categories (e.g., whether material precedes or follows age, or whether origin precedes or follows opinion). This requires models to reason with partial information and avoid relying on prior knowledge. 

We observe an interesting failure mode in this task: some models correctly induce partial adjective orderings but rely on prior knowledge of standard English order when selecting answers. Consider the example below.

\begin{tcolorbox}[colback=black!5!white,colframe=black!75!black,title=Prompt]
In a variant of English, we are given that the following sentences have correct adjective order: (1) medium-size aqua wrench (2) normal-size spherical repulsive scissors $\cdots$ (90) big chair (91) midsize bicycle \\\\
In this variant of English, which of the following sentences use the correct adjective order? $\cdots$
Select all that apply.\\
(A) big yellow drinking prismlike mysterious bangladeshi fiberglass brand-new ball\\
(B) old leather drinking pyramidal lovely congolese teal key\\
$\cdots$\\
(J) ancient paper smoking circular repulsive turkish bottle\\
(K) None of the above
\end{tcolorbox}

o3-mini (high) provides the following response: 

\begin{tcolorbox}[colback=black!5!white,colframe=black!75!black,title=Model Response]
The answer is: ACDEHI

In every correct sentence the adjectives from chain A (if present) appear in the order: size → color → purpose → shape → opinion → origin; and {\color{red}if any chain‑B adjectives are used they come afterward in the order}: material → age.\\\\
Looking at the options:\\
(A) “big (size) yellow (color) drinking (purpose) prismlike (shape) mysterious (opinion) bangladeshi (origin) fiberglass (material) brand‐new (age) ball” fits exactly.\\
(C) “large (size) gray (color) whittling (purpose) spherical (shape) silly (opinion) turkish (origin) brand‐new (age) banana” fits (chain‑A adjectives then the single chain‑B adjective).\\
(D) “medium-size (size) brown (color) walking (purpose) triangular (shape) ridiculous (opinion) russian (origin) dog” {\bf{fits chain A perfectly}}.\\
(E) “little (size) crimson (color) driving (purpose) circular (shape) ridiculous (opinion) congolese (origin) rubber (material) bowl” fits (chain‑A then chain‑B).\\
(H) “red (color) typing (purpose) circular (shape) good (opinion) vietnamese (origin) paper (material) archaic (age) candle” fits.\\
(I) “typing (purpose) pyramidal (shape) obnoxious (opinion) mexican (origin) cardboard (material) brand‐new (age) bag” fits as well.

\end{tcolorbox}

Although the model correctly infers the partial adjective orderings: (i) size → color → purpose → shape → opinion → origin, and (ii) material → age, it incorrectly assumes that {\color{red}material follows origin}. This assumption likely stems from prior knowledge of standard English ordering (opinion → size → age → shape → color → {\color{red}origin → material} → purpose). While the model correctly identifies option D, the only correct answer, its incorrect assumption leads it to also select options A, C, E, H, and I.

\subsection{Linguini}
This task comes from \citet{sanchez2024linguini} and its problems are extracted from the International Linguistic Olympiad (IOL). According to the original work that introduced this dataset, the problems are \emph{"linguistic problems which require meta-linguistic awareness and deductive reasoning capabilities to be solved instead of pre-existing language proficiency"}. 

We created a subset of the Linguini problems by sampling from four categories of the Linguini problems, namely \emph{translation}, \emph{fill blanks}, \emph{num to text} and \emph{text to num}. The original dataset contains questions that require multiple answers. For example, the \emph{fill blanks} questions have multiple blanks that need to be filled. We create questions that have a single answer by randomly selecting one of those blanks and only asking the model to fill that one.

\subsection{Movie Recommendation}
The original Movie Recommendation task in BBH has been created as follows. For each question, a set of eight movies from MovieLens have been selected such that a rather large number of people have all liked five of them and disliked three of them. Then, a question has been generated by giving four of the five liked movies and asking models to recommend one of the remaining four movies, where the correct answer is the one left out of the 5 liked movies. 

We updated this task as follows. We create multiple sets of movies where one of them contains the five liked movies and the other ones contain some of the liked movies and some of the disliked movies. Then, we ask the model to select the set that contains movies that are more likely to all be liked by a large group of people. In the new variant we created, instead of recommending a single movie given four movies, models have to examine each set separately and predict their overall likability, and then decide the option that is more likely to have a likability score with our specific definition of likeability.

\subsection{Multi-step Arithmetic}
This task introduces new arithmetic operators. An example of such an operator is as follows:
\begin{tcolorbox}[colback=black!5!white,colframe=black!75!black,title=Sample New Operator]
a >< b equals (a - b) if a * b > 0; otherwise, it equals a + b
\end{tcolorbox}
Some of the operations can be defined based on the other new operations. For example we may have:
\begin{tcolorbox}[colback=black!5!white,colframe=black!75!black,title=Another Sample New Operator]
a ; b equals (a >< b) if a + b > 0; otherwise, it equals a - b
\end{tcolorbox}
We also define a form of composing multiple operations as follows: a <op1><op2> b denotes (a op1 b) op2 b; for example,  4 +* -5 means (4 +~ 5) * -5 and 4 *++ 5 means (4 * 5) ++ 5. 

Then we sample random arithmetic expressions involving the above operations. An example expression is: (1 @*+ 4) <>+[] (-4 *<>* -1) (although our expressions are longer), with @, <>, and [] being new operations. The job of the model is to compute the value of the expression. Being able to compute these expressions requires expanding the expressions and making a long list of computations correctly.

Upon looking at the outputs generated by the models, we find a common failure mode is that when multiple operations are composed, models sometimes forget to apply all of them despite understanding how the operator composition works. For instance, in one of our examples, while o3-mini correctly explains how the operator composition works in its reasoning trace, it still computes $(1\quad * > < - \quad-6)$ as $(1 * -6)\quad ><\quad-6$ and forgets the final subtraction operator in the composed operator $* > < -$. 

\subsection{NYCC}

This task builds on the existing benchmarks for the New Yorker Caption Contest (NYCC) dataset \cite{hessel2022androids, zhang2024humoraimassivescale}. The NYCC caption dataset consists of a) several hundred contests, each of which is a cartoon published in the New Yorker magazine and several thousand submitted humorous captions, b) crowdsourced ratings for each caption. The ratings are on a scale of ``Unfunny'', ``Somewhat Funny'', and ``Funny'', and each caption has anywhere from a few dozen to a few thousand ratings.  Past works have focused on pairwise comparison tasks, where two captions and a textual description of the cartoon are presented to the model, and the model has to pick the funnier of the two. As discussed in these works, the model tends to be fairly successful at these tasks, with GPT-4 Turbo getting to $\approx$ 70\% accuracy. 

To make the task significantly more difficult, for each contest we sample one query from the top ten rated, and then take captions ranked 1000-1009 and ask the model to choose the funniest. We use the ``Canny'' textual descriptions of the cartoons generated by GPT-4o that are provided in \citet{zhang2024humoraimassivescale}. An example query is below.

\begin{tcolorbox}[colback=black!5!white,colframe=black!75!black,title=Sample Problem]
The following is a description of a funny cartoon for the New Yorker Caption Contest

  Description: Two people are sitting at a table in a restaurant, having a conversation over glasses of wine. One of them is wearing a suit of armor.

Which of the following captions is the funniest?\\
1) Yes, I wrote that in my profile but I didn’t mean it literally.\\
2) That's not what I meant by a nice night out and you know it.\\
3) Sorry. It’s laundry day.\\
4) So, do you like horses?\\
5) In vino veritas? Surely you joust.\\
6) So that’s your best suit?\\
7) ``Cougar seeks millennial'' didn’t mean the year 1000\\
8) Oh, really? You think men are under attack?\\
9) This is not what I expected when you said you were middle aged.\\
10) Frankly, you look much older than your profile photo
\end{tcolorbox}

In the above the correct caption is caption number 9. Adding multiple possible options makes the task significantly more challenging compared to the pairwise task.

\subsection{Object Properties}
In this task, an initial collection of objects with different properties (color, size, origin, smell, and material) are provided (e.g., a extra-small blue Canadian jar made of glass and with a smell of rose). Then, the collection goes through several updates corresponding to adding, removing or editing some of the objects. The updates are explained in the prompt and the models require a full grasp of the object properties to identify what changes to the collection must be made for each update. A simple example of an update is as follows: 

\begin{tcolorbox}[colback=black!5!white,colframe=black!75!black,title=Sample Update to the Collection]
My dad threw away all objects of a certain color from my collection.\\After this, my collection only had 5 blue objects and 3 white objects.
\end{tcolorbox}

For the above update, one has to find which color has been removed by comparing the new colors with the object colors in the previous collection, and then update the collection accordingly. The set of updates that the collection goes through in each of the examples are randomly selected from a large set of possible changes. At the end, a question is asked about the final collection. The question is either an \emph{either} question in which we ask how many items in the final collection have property 1 or property 2, ... (e.g., how many items are either blue or small), or a \emph{neither} question in which we ask how many items neither have property 1 nor property 2, ... (e.g., how many items are not blue and are not small).

One of the updates, in particular, is a trick update. It says one of the objects of property X has been lost, but does not say which one. For example it says: \texttt{I lost one of the blue objects} without specifying which blue object. This update is made in a way that the final number is the same regardless of which object was lost. However, we observe that many of the models get confused with this update and assume that the final count cannot be computed since we do not know which item was lost. Specifically, o3-mini (high) says the problem cannot be solved in 8\% of such cases, Gemini 2.0 Flash in 93.5\% of the cases, and GPT4o in 96.5\% of the cases.

\begin{table}[t]
\centering
\footnotesize
\begin{tabular}{c|ccccc}
\toprule
\textbf{Models} & Gemini 2.0 Flash-Lite & Gemini 2.0 Flash & GPT4o & DeepSeek R1 & o3-mini (high) \\
\midrule
\textbf{Over-count \%} & 53.0 & 50.5 & 17.5 & 6.0 & 1.5 \\
\textbf{Under-count \%} & 43.0 & 38.5 & 66.5 & 15.5 & 8.5 \\
\bottomrule
\end{tabular}
\caption{Percentage of problems from the Object Counting task where the models over-counted or under-counted.}
\label{tab:object-counting}
\end{table}

\subsection{Object Counting}
Given a long list of objects that a person has, the model has to count the number of items of a certain type. For examples, the items might belong to classes (fruits, cell phones, cars) and the goal may be to count the total number of cell phones that the person has. We consider two types of questions: 1- counting the sum of the number of items belonging to two different classes, and 2- finding the absolute difference of the number of items belonging to two different classes. To add to the difficulty of the task, some irrelevant information, including the number of the same items that other people have, are added to the input context so the problem becomes one of finding multiple needles in a haystack.

In Table~\ref{tab:object-counting}, we report the percentage of cases where each of the models either over-counted the number of objects or under-counted, for the subset where the sum of two sets must be reported. Interestingly, we observe a that different models have different failure modes on this task. The Gemini models tend to mostly over-count when they are wrong, whereas GPT4o, DeepSeek R1 and o3-mini tend to under-count when they are wrong.

\subsection{SARC Triples}
SARC (Self-Annotated Corpus for Sarcasm) \citep{khodak2017large} is a large dataset of sarcasm responses mined from the Reddit social media / forum platform. Many Reddit users end a post or reply with the token ``\texttt{/s}'' when they have intended the preceding text to be interpreted sarcastically or satirically. This allowed positive examples of user-intended sarcasm to be mined.

Forking off the SARC dataset, we construct a challenging task for LLMs that requires reading three  independent examples from SARC, and classifying each into binary label, where a positive label indicates sarcasm. The SARC authors created a balanced test set with 64,666 examples. Many of these examples can only be understood with an image or an article link that accompanied the original post or reply. On the other hand, some examples, usually with longer textual content, can be understood on their own. We design our derived benchmark to consist mainly of the latter type. To achieve this, we filter out examples with either (1) less than 100 characters or (2) without a reply, resulting in 679 examples from the original test set, with 48.4\% positive label rate. We sample (uniformly-at-random) 600 examples from this set, group them (uniformly-at-random) into groups of three, and pass the text of each 3-tuple of post, reply pair to the following prompt:

\begin{tcolorbox}[colback=black!5!white,colframe=black!75!black,title=Prompt Template]
Here are three (post, reply) pairs from Reddit. Your task is to decide whether\\each reply is sarcastic. Specifically, label each pair with a "0" or "1", where\\a "1" indicates that the reply is sarcastic, and a "0" indicates that the reply\\does not contain sarcasm, and provide your final answer as a comma-separated set of labels (e.g., "1,0,0" or "0,0,0").\\POST 1: post1\_text\\REPLY 1: reply1\_text\\POST 2: post2\_text\\REPLY 2: reply2\_text\\POST 3: post3\_text\\REPLY 3: reply3\_text
\end{tcolorbox}

\subsection{Shuffled Objects}
The original task in BBH is as follows: there are N people each assigned to an object/person (e.g., a dance partner, a book, a color, etc.). For example, Alice has a green book, Bob has a red book, etc.  Then, there are multiple switch operations where pairs of people switch together what they are assigned to (e.g., Alice and Bob switch their books). At the end, one needs to predict the object/person assigned to one of the N people (e.g., at the end, what color is the book that Bob has?).

We created two variants of this problem. In the first variant, we keep everything the same except that we add switch actions that have no effect. For example, we add \texttt{Then, Person1 and Person2 switch their books. Then, Person2 and Person1 switch their books}. We add many of these no-effect operations so that the problem becomes a long-context reasoning problem similar to the approach in \citet{vodrahalli2024michelangelo}. 

The second variant extends the first variant, in which we assign names to some of the switch actions as they occur and use those names later. For example, the first time \texttt{Person1 switches with Person2} occurs, we replace the text with \texttt{Person1 switches with Person2 (let's call this Action K)}, and the next time the same switch happens, with some probability we replace the text with \texttt{action K repeats}. Given the long-context nature of the problem, the model requires to have the ability to remember information from many steps ago to be able to identify what that action corresponded to.

A naive approach to the problem in these tasks is to look at the switch operations one by one and keep updating the object/person assigned to each of the N people. This, however, will require track keeping over a very large number of operations. While reasoning-specialized models might still be able to do this thanks to their long outputs, this may be less feasible for the general-purpose models, and it is not the optimal solution to the problem. A more clever approach to the problem is to first identify all the operations that cancel each other out, and then do the track keeping only over the few operations that do not cancel. 

Looking at some model traces, we observe that the models typically adopt the non-optimal approach of updating the state after each switch, thus running out of output tokens in some cases. For example, Gemini 2.0 Flash runs out of output tokens for 25\% of the problems. We also observe a second failure case where, if the question asks about the person/object assigned to Person P at the end, the model assumes only switches involving Person P are important and other switches are irrelevant. This is, however, not True. To understand why, suppose "A" is assigned to "a", "B" is assigned to "b", and "C" is assigned to "c". Then "A" and "B" switch and then "B" and "C" switch, and then we want to know what "C" is assigned to. If we only consider the switches involving "C", then we may predict the correct answer to be "b" whereas the correct answer in this case is "a".

\subsection{Spatial Reasoning}
This task is mainly based on the problems in SpacialLLMEval \citep{yamada2023evaluating}. The problems describe a geometric construct composed of vertices and edges. At each vertex, there is a unique object. An agent starts from one of the vertices, moves along the edges and observes the objects at several vertices, and then after moving for several steps along the edges, the job of the model is to determine what object is at the final vertex where the agent stops.

We sampled from the hexagonal, circular, and rhombus constructs of SpacialLLMEval. We also created similar constructs with tree structure, triangular and diamond shapes and increased the difficulty compared to the problems in SpacialLLMEval by increasing the number of hops of reasoning (corresponding to the number of moves of the agent). Moreover, while the original problems and the aforementioned problems we created mainly require keeping track of the state after each move, we also create some variants of the problem where we provide multiple paths that intersect at some vertex, thus requiring backward reasoning from the intersection point to identify the position of the previous objects. As an example, consider the problem below:

\begin{tcolorbox}[colback=black!5!white,colframe=black!75!black,title=Sample Problem Requiring Backward Reasoning]
You have been given a diamond tile map consisting of N rows [...] There is a unique object placed at each vertex. [...] You are initially at the top corner where you see a football. Then you move down-right for one step and see a shampoo. Then you move down-left for one step and you see a cat. [...] Then, you jump to a random vertex V where you see a bear. Then you move [...] Then you move up-left and you see a shampoo. Then you jump back to the random vertex V and do the following moves: down-left, down-left, down-right, up-left, down-left, up-left. What will you find?
\end{tcolorbox}

For the first path (up until the first random jump), we know where the path starts and we can use that along with the following moves to determine which object is where. Then, a random jump is made to a vertex V but it is not specified which vertex it is. However, we observe that after a number of moves, the agent sees the \emph{shampoo} again so it can reason backward from this point to figure out which vertex it has been at in the previous steps. These information can also be used to determine the vertex V which must then be used to solve the problem when the second jump to V is made.

We find that the problems involving backward reasoning are more challenging for the models. Specifically, we find that o3-mini (high) gives an accuracy of 58.8\% on the forward-only problems and 19.2\% on the backward problems and DeepSeek R1 gives an accuracy of 48.6\% on the forward-only problems and 3.8\% on the backward problems. 

\subsection{SportQA}
SportQA \citep{xia2024sportqa} is a challenging sports understanding dataset designed to test rule-based and strategic reasoning capabilities in LLMs beyond surface-level sports knowledge. It consists of three levels (Level 1 to Level 3) with increasing difficulty. In this work, we focus on Level 3 questions, which are curated by coaches and student athletes across six sports: soccer, basketball, volleyball, tennis, table tennis, and American football. We sub-sample 200 multi-hop reasoning questions and discard single-hop questions from the Level 3 set. Overall, these questions challenge LLMs to reason about fine-grained sports rules (e.g., penalty assessment and tactical choices), which expert student athletes can answer with near-perfect accuracy \citep{yang2024sports}.

The questions we selected from SportQA have a compositional nature, in that a main question and some sub-questions are provided and the model has to answer all of them correctly for its answer to be considered correct. We add some instructions at the end of the questions so the models can answer them zero-shot in the format that we want. The questions are in the following format:

\begin{tcolorbox}[colback=black!5!white,colframe=black!75!black,title=Prompt Template]
You will be given a main question and two sub-questions. Each question comes with multiple choices you can select from. For each question, select all the correct choices.\\
Main Question: <TEXT OF THE QUESTION>\\
<OPTIONS A to D>\\
Sub-Question 1: <TEXT OF THE SUB-QUESTION>\\
<OPTIONS A to D>\\
Sub-Question 2: <TEXT OF THE SUB-QUESTION>\\
<OPTIONS A to D>\\
For each question, provide the answer as a concatenation of the correct choices. Separate the answers for the questions by comma. For example, if the correct choices for the main question are A and C, for the first sub-question is D and for the second sub-question are B and D, your final answer must be "AC, D, BD".
\end{tcolorbox}

\subsection{Temporal Sequences}
In this task, the calendar schedules of a few people is provided for an entire week. The blocked times for the calendar of each person is sampled randomly, and is provided as text either by giving the times of the day when it is blocked or giving the times of the day when it is free. The goal is to find: 1- the longest meeting that can be scheduled for them, and 2- the number of possibilities for such a meeting. These people may also have some constraints or we might have some information about them that has to be taken into account for meeting scheduling. Examples include: being in a different timezone than the other participants, needing some free time before/after the meeting, being flexible to miss a portion of the meeting, requiring some free time for lunch, only being able to attend meetings of up to a certain length, being willing to free up some specific parts of the day if needed, etc.

The model predictions are considered correct if they predict both values (i.e. the longest time and the number of possibilities for a meeting of that length) correctly. We observe that both of these add to the difficulty of the problem. For example, o3-mini (high) has an overall accuracy of 68.5\% on this problem, but if we only asked for the longest meeting time, then the accuracy will jump to 78\%. The same is true for Gemini 2.0 Flash and Flash-Lite where their accuracy for the overall task is respectively 0.5\% and 1.0\%, but if we had asked only for the longest meeting time then their accuracy will jump to 5\% and 7\% respectively.

\subsection{Time Arithmetic}
This task is based on the time arithmetic subset of the Test of Time (ToT) benchmark \citep{fatemi2024test}. The original subset contains various questions about understanding, computations over, comparisons, and conversions of dates and times. There are also trick questions which may require extra thinking. The dataset also contains some scheduling problems, but we removed that subset given that we have an entire task (Temporal Sequence) dedicated to it.

Following \citet{hosseini2024not}, we created a compositional version of the ToT Time Arithmetic dataset as follows. Let Q1 and Q2 be two questions from the original dataset, where the answer to the Q1 is A1 (A1 being a number) and let A2 be a number that is used in Q2. Then, we create a compositional question as follows:

\begin{tcolorbox}[colback=black!5!white,colframe=black!75!black,title=Sample Prompt Template]
Let the answer to Q1 be X.\\Q1: <Text of the question>.\\Let X' = X + (A2 - A1). Use this value to solve Q2.\\Q2: <Text of the question with A2 replaced with "X'">.
\end{tcolorbox}

In some cases, the answer to a question might contain multiple numbers, e.g. a date with three numbers. In those cases, we assign these values to variables X, Y, and Z and use them in the later questions.

\subsection{Web of Lies}
In this task, whether a specific person P1 tells the truth of lies is provided as input. Then, for other people, it is specified what they say about the truth value of some other person. This forms a chain-like structure that can be started from P1 and continued to find whether each of the people tells the truth or lies.

We used two different variants for this task. The first variant comes from the \emph{web of lies V2} from LiveBench \cite{white2024livebench}. In this variant, complexity has been added to the task by specifying where each person is, and then having sentences such as \emph{The person at the cafe says the person at the zoo lies}. The second version is created by us. In this version, we add cyclic cases whose truth value remains unknown, but one can still infer something about them and continue the chain. For example, consider a cyclic case such as \emph{Person1 says Person2 tells the truth. Person2 says Person1 tells the truth.} In this case, we cannot determine whether Person1 or Person2 tell the truth or lie (so their truthfulness remains unknown). However, if we have another sentence \emph{Person3 says either both Person1 and Person2 lie or both tell the truth}, we can determine that Person3 tells the truth. In both variants of the problems, we ask about the truthfulness of three of the people in the chain, so the random chance performance for the LiveBench subset is $1/8$ since the truthfulness of each of the three people can be either yes or no, and $1/27$ for our new set given that the values can also be unknown.

We observe that the first subset is easier than the second subset, so we included only 40 examples of subset one and 160 of subset 2. Specifically, o3-mini (high) gives an accuracy of 100\% on subset one and Gemini 2.0 Flash gives an accuracy of 77.5\%, whereas on the second subset the two models give an accuracy of 28.8\% and 3.8\%. We still keep the first subset despite the high performance of the o3-mini model so it can be used to distinguish among smaller, general-purpose models.

For the new set we created, we always ask about the truthfulness of one of the people at the end of the chain, and another person at an earlier position in the chain (but still far off in the chain). Conceptually, one would expect that if a model has made a mistake for the person at an earlier position in the chain, then the chances of making a mistake for the person at the end of the chain must be higher. We verified whether this is the case for our models. For o3-mini, we observe that the accuracy for both cases is 41.2\%, for Gemini 2.0 Flash it is 30\% for the earlier person and 27.5\% for the last person, and for Gemini 2.0 Flash-Lite it is 25.6\% for the earlier person and 21.2\% for the last person, all showing this effect. GPT4o, however, is surprisingly behaving the opposite, having an accuracy of 19.4\% for the earlier person and 25\% for the last person.

\subsection{Word Sorting}
The Word Sorting task is split into 2 sub-tasks.

The first sub-task is from the BIG-Bench Mistake dataset \citep{tyen-etal-2024-llms}. This task involves finding the first mistake in an existing chain-of-thought sequence, used to answer a Word Sorting question in the original BBH dataset. In each example, the target answer is either the number where the first mistake occurred, or that there are no mistakes in the CoT sequence. These CoT sequences are generated by prompting PaLM 2 Unicorn \citep{anil2023palm2technicalreport} on the original BBH dataset at temperature = 0.
The newline is used as a stop token so that each intermediate step can be prepended with `Thought 1: ', `Thought 2: ', etc. Further information on the prompting and generation process can be found in \citet{tyen-etal-2024-llms}.

The second sub-task is sorting a list of words given a new alphabet order (examples include: an alphabet order that is the same as English but two letters are swapped in the order, an alphabet order that is the same as English but one/two letters are moved to the beginning/end of the order, or a completely new order). This task requires going against a strong prior and sorting words in a non-typical way. We observe an interesting failure mode for this task where some models understand the new alphabet order correctly but keep sorting the words based on the original English order. As an example, consider the following question:

\begin{tcolorbox}[colback=black!5!white,colframe=black!75!black,title=Prompt]
Consider a new alphabet whose letters have the same order as the English alphabet, except that s and i are the last two letters. Sort the following words with the new alphabet and separate them with comma: va, sprue, subject, immemorial, anaheim, interpretations, clinic, trafficked, communications, marx, demonstrations, micro, offprint, characterizations, personifications, eaten, madeira
\end{tcolorbox}

Gemini 2.0 Flash provides the following response:

\begin{tcolorbox}[colback=black!5!white,colframe=black!75!black,title=Model Response]
Here's how to sort the words according to the modified alphabet: 1. **Modified Alphabet:** a, b, c, d, e, f, g, h, j, k, l, m, n, o, p, q, r, t, u, v, w, x, y, z, s, i [...]
The final answer is: anaheim, characterizations, clinic, communications, demonstrations, eaten, immemorial, interpretations, madeira, marx, micro, offprint, personifications, sprue, subject, trafficked, va
\end{tcolorbox}

While the model provides the correct new alphabet, we see that the final answer is sorted based on the original order (which agrees better with model's prior), and not based on the new order.

\begin{table}[t]
\centering
\footnotesize
\begin{tabular}{c|cccc}
\toprule
\textbf{Puzzle Size / Models} & Gemini 2.0 Flash-Lite & Gemini 2.0 Flash & GPT4o & o3-mini (high) \\
\toprule
\textbf{5 x 5} (with distractors) & 37.5 & 60.0 & 37.5 & 90.0 \\
\textbf{6 x 6} (with distractors) & 38.9 & 33.3 & 30.6 & 58.3 \\
\textbf{7 x 7} (with distractors) & 36.5 & 36.5 & 37.8 & 55.4 \\
\textbf{8 x 8} (without distractors) & 16.0 & 52.0 & 20.0 & 74.0 \\
\bottomrule
\end{tabular}
\caption{Accuracy break-down of the model performances on the Zebra Puzzles task by puzzle size.}
\label{tab:zebra-puzzles}
\end{table}

\subsection{Zebra Puzzles}
Zebra puzzles, also known as Einstein puzzles, are verbal descriptions of entities and properties that partially populate a grid linking entities to their properties \citep{shah2024causal}. The description may also include constraints on these properties, such that it is possible to deduce the other entity-property links. Following the approach in \citet{shah2024causal}, we generate square-grid Zebra puzzles of size 5x5, 6x6, 7x7, and 8x8. We add distracting clues to puzzles of size 5, 6, and 7 to make them more challenging, but do not add them to puzzles of size 8 to avoid keeping the context size too large. To simplify evaluation, the questions ask for the position of one of the n people in the n x n puzzles, so the random chance performance for a n x n puzzle is $1 / n$.

Looking at the performance for the top performing models in Table~\ref{tab:zebra-puzzles}, we notice two interesting patterns. Firstly, while the random chance performance is lower for larger puzzles, we see that increasing the puzzle sizes does not significantly lower the model performances in some cases. This is especially true when going from puzzles of size 6 to 7. Secondly, we observe that the top-2 models (Gemini 2.0 Flash and o3-mini high) perform significantly better on the 8x8 puzzles that have no distracting clues compared to the smaller puzzles of size 7x7 or 6x6 that do have distracting clues. This hints at a possible failure mode for frontier models: they seem to get confused in presence of distractors and underperform. This might be in part due to the fact that models might have already seen clean zebra puzzles and their solutions, but they may not have seen the variant with distracting clues, making the latter a case of out-of-distribution generalization.

\section{BBEH vs BBH Performance}
To understand how much each task in BBEH has become harder compared to its counterpart in BBH, we evaluated Gemini 2.0 Flash on BBH and reported the results in Table~\ref{tab:comp-results}. For the fairness of the comparison, we ran the model in a zero-shot setting. However, we note that some of the tasks in BBH may become slightly ambiguous in a zero-shot setting given that it has been mostly developed for a few-shot evaluation. Nevertheless, we observe that on almost all the tasks, the difficulty level has significantly increased in BBEH. A notable exception is the DisambiguationQA task. Checking the responses from Gemini 2.0 Flash on the BBH version of the dataset, we observe that the model overly selects the ambiguous option, sometimes for potentially legit reasons. For example, for disambiguating the pronoun \emph{they} in the sentence \emph{Alex told us that they could not meet}, the model responds that \emph{They could refer to Alex or to some other group of people not explicitly mentioned. Therefore, the antecedent is ambiguous}. We also find that in several of the cases, simply changing the task description from \emph{explain the antecedent of the pronoun [...] or state that it is ambiguous} to \emph{try to disambiguate the antecedent of the pronoun given the context or state that it is ambiguous if it cannot be disambiguated} makes the model pick the right choice.

\begin{table}[th!]
\centering
\begin{tabular}{c|cc}
\textbf{Task in BBEH $\downarrow$ / Accuracy on $\rightarrow$} & Old task from \textbf{BBH} & New task in \textbf{BBEH} \\
\toprule
\textbf{BoardgameQA} & 88.0 & 42.5 \\
\textbf{Boolean Expressions} & 97.6 & 27.0 \\
\textbf{Causal Understanding} & 65.2 & 52.0 \\
\textbf{DisambiguationQA} & 42.0 & 48.3 \\
\textbf{Dyck Languages} & 65.2 & 14.0 \\
\textbf{Geometric Shapes} & 73.6 & 35.0 \\
\textbf{Hyperbaton} & 94.8 & 4.5 \\
\textbf{SARC Triples} & 86.0 & 37.5 \\
\textbf{Linguini} & 62.8 & 15.5 \\
\textbf{Movie Recommendation} & 66.4 & 59.5 \\
\textbf{Multistep Arithmetic} & 99.6 & 9.5 \\
\textbf{NYCC} & 81.2 & 11.0 \\
\textbf{Object Properties} & 96.8 & 1.5 \\
\textbf{Object Counting} & 97.6 & 11.0 \\
\textbf{Shuffled Objects} & 100.0 & 9.0 \\
\textbf{Spatial Reasoning} & 97.6 & 18.5 \\
\textbf{SportQA} & 89.6 & 23.0 \\
\textbf{Buggy Tables} & 98.6 & 3.5 \\
\textbf{Temporal Sequences} & 98.8 & 0.5 \\
\textbf{Time Arithmetic} & 92.0 & 48.0 \\
\textbf{Web of Lies} & 94.8 & 18.5 \\
\textbf{Word Sorting} & 84.8 & 26.0 \\
\textbf{Zebra Puzzles} & 87.6 & 44.5 \\
\toprule
\textbf{BBEH} & 85.2 & 23.90 \\
\end{tabular}
\caption{Performance of Gemini 2.0 Flash on BBEH and its counterpart task from BBH.}
\label{tab:comp-results}
\end{table}

\section{Reproducibility}
For most of the models we tested in this work, we obtained the results through API calls. Specifically, for Gemini 2.0 and Gemma2 models, we used AI Studio, for GPT4o and o3-mini (high) we used the OpenAI API, and for DeepSeek R1 results we used the API from Together AI. For the LLama and Distill R1 Qwen results, whose parameters are available publicly, we obtained results by loading them on GPUs. This ensures reproducibility as other researchers can also follow the same approach and obtain our results.

For all of our problems, we added the following suffix to the text of the question to encourage the model to produce the final answer in a format that we can easily extract it:

\begin{tcolorbox}[colback=black!5!white,colframe=black!75!black,title=Suffix]
Think step by step, and when you provide the final answer, please use the prefix "The answer is:" without any modification, and provide the answer directly, with no formatting, no bolding, and no markup. For instance: "The answer is: 42" or  "The answer is: yes". If the question is multiple choice with a single correct answer, the final answer must only be the letter corresponding to the correct answer. For example, "The answer is: (a)".
\end{tcolorbox}

We then extract the answer by splitting with the prefix and finding what comes next. We find that sometimes the models slightly deviate from the exact prefix we gave them, so we look for four prefixes in the answer until one of them is found: \texttt{"The answer is: ", "The answer is ", "The final answer is: ", "The final answer is "}. Once we extract the final answer, we apply some minimal cleaning as follows: 1- if the final answer is wrapped within the \emph{boxed}, \emph{text}, \emph{texttt} or \emph{**}, we remove that and extract what is inside it. We notice that after producing the final answer, some models produce a "\texttt{\textbackslash n}" and then some extra text. Therefore, we split the extracted final answer using "\texttt{\textbackslash n}" and take the first element as the final answer. Then we lowercase both the final answer and the label determine correctness using a few simple rules: 1- if the two are identical, then we consider the final answer correct, 2- if the final answer is identical to the label up to removing single or double quotes or brackets from the beginning and end of it, we consider it to be correct, 3- the label for multi-choice questions is in (<LETTER>) format and we expect a similar final answer but if the final answer is only the letter without the parentheses, we consider it correct, and finally 4- for questions whose labels contain multiple elements separated by comma, if the label and final answer are the same up to replacing the spaces after the commas with blanks, then we consider the final answer to be correct.

\end{document}